\pdfoutput=1

\documentclass[11pt]{article}

\newif\ifreview

\ifreview
\usepackage[review]{emnlp2021}
\else
\usepackage{emnlp2021}
\fi

\usepackage{times}
\usepackage{latexsym}

\usepackage[T1]{fontenc}

\usepackage[utf8]{inputenc}

\usepackage{microtype}
\usepackage{soul}
\usepackage{booktabs}
\usepackage{graphicx}
\usepackage{makecell}
\usepackage{subfig}
\usepackage{balance}

\newcommand{\M}{\textsc{massive}}
\graphicspath{ {./img/} }

\let\svthefootnote\thefootnote
\newcommand\freefootnote[1]{%
  \let\thefootnote\relax%
  \footnotetext{#1}%
  \let\thefootnote\svthefootnote%
}

\title{MASSIVE: A 1M-Example Multilingual Natural Language Understanding Dataset with 51 Typologically-Diverse Languages}

\ifreview
\author{Jack FitzGerald \\\And
  Christopher Hench \\\And
  Charith Peris \\\AND
  Scott Mackie \\\And
  Kay Rottmann \\\And
  Ana Sanchez \\\AND
  Aaron Nash \\\And
  Liam Urbach \\\And
  Vishesh Kakarala \\\AND
  Richa Singh \\\And
  Swetha Ranganath \\\And
  Laurie Crist \\\AND
  Misha Britan \\\And
  Wouter Leeuwis \\\And
  Gokhan Tur \\\AND
  Prem Natarajan \\}
\else
\author{Jack FitzGerald\footnote{Corresponding author, \texttt{jgmf@amazon.com}. All authors were associated with Amazon at the time of publication.} \\\And
  Christopher Hench \\\And
  Charith Peris \\\AND
  Scott Mackie \\\And
  Kay Rottmann \\\And
  Ana Sanchez \\\AND
  Aaron Nash \\\And
  Liam Urbach \\\And
  Vishesh Kakarala \\\AND
  Richa Singh \\\And
  Swetha Ranganath \\\And
  Laurie Crist \\\AND
  Misha Britan \\\And
  Wouter Leeuwis \\\And
  Gokhan Tur \\\AND
  Prem Natarajan \\}
 \fi
  
\begin{document}
\maketitle


\begin{abstract}
\ifreview
\else
We present the MASSIVE dataset---Multilingual Amazon Slu resource package (SLURP) for Slot-filling, Intent  classification, and Virtual assistant Evaluation.
\fi
MASSIVE contains 1M realistic, parallel, labeled virtual assistant utterances spanning 51 languages, 18 domains, 60 intents, and 55 slots. MASSIVE was created by tasking professional translators to localize the English-only SLURP dataset into 50 typologically diverse languages from 29 genera. We also present modeling results on XLM-R and mT5, including exact match accuracy, intent classification accuracy, and slot-filling F1 score. We have released our dataset, modeling code, and models publicly.
\end{abstract}

\ifreview
\else
\freefootnote{*Corresponding author, \texttt{jgmf@amazon.com}. All authors were associated with Amazon at the time of publication.}
\fi

\section{Introduction and Description}

Natural Language Understanding (NLU) is a machine's ability to understand the meaning and relevant entities from text.
For instance, given the utterance \texttt{what is the temperature in new york}, an NLU model might classify the intent as \texttt{weather\_query} and fill the slots as \texttt{weather\_descriptor: temperature} and \texttt{place\_name: new york}.
Our particular focus of NLU is one component of Spoken Language Understanding (SLU), in which raw audio is first converted to text before NLU is performed \citep{Young2002TalkingTM, Wang2005SpokenLU, Tur2011SpokenLU}.
SLU is the foundation of voice-based virtual assistants like Alexa, Siri, and Google Assistant.
Though virtual assistants have advanced incredibly in the past decade, they still only support a small fraction of the world's 7,000+ languages \citep{ethnologue}.
Challenges for multilingualism span the software stack and a variety of operational considerations, but one difficulty in creating massively multilingual NLU models is the lack of labeled data for training and evaluation, particularly data that is realistic for the task and that is natural for each given language.
High naturalness typically requires human-based vetting, which is often costly.


\ifreview
\else
We present \M{} (\textit{M}ultilingual \textit{A}mazon \textit{S}LU Resource Package (SLURP) for \textit{S}lot filling, \textit{I}ntent classification, and \textit{V}irtual assistant \textit{E}valuation), a new 1M-example dataset composed of realistic, human-created virtual assistant utterance text spanning 51 languages, 60 intents, 55 slot types, and 18 domains.
With the English seed data included, there are 587k train utterances, 104k dev utterances, 152k test utterances, and 153k utterances currently held out for the MMNLU-22 competition, which will be released after the competition.
We have released our data, code, and models \footnote{\href{https://github.com/alexa/massive}{https://github.com/alexa/massive}}.
\fi

\M{} was created by localizing the SLURP NLU dataset (created only in English) in a parallel manner. SLURP is described further in Section \ref{sect:related}, linguistic analyses of the dataset in Section \ref{sect:linguistics}, and the localization process in Section \ref{sect:localization-workflow}. Results for Massively Multilingual NLU (MMNLU) modeling, in which a single model can perform NLU on any of the incoming languages, are given in Section \ref{sect:modeling}.

\section{Related Work} \label{sect:related}

Prior researchers have emphasized the need to explore the unique challenges of low-resource languages \citep{simpson2008human, strassel-tracey-2016-lorelei, cruz2020establishing, lakew2020low, marivate-etal-2020-investigating, magueresse2020lowresource, goyal2021flores101}, while the growing number and size of language models (mBERT \citep{devlinmBERT}, RoBERTa \citep{liu2019roberta}, XLM \citep{lample2019crosslingual}, XLM-R \citep{conneau-etal-2020-unsupervised}, mBART \citep{liu2020multilingual}, MARGE \citep{lewis2020pretraining}, and mT5 \citep{xue-etal-2021-mt5} pre-trained on massively multilingual corpora have allowed for significant improvements in supporting them.
However, the creation of evaluation datasets for specific tasks has not kept pace.
Some tasks,  such as Named Entity Recognition (NER) or translation, lend themselves to mining existing corpora \citep{TIEDEMANN12.463,Pan2017CrosslingualNT,hu2020xtreme}, while others such as NLU, the focus here, require the creation of new data and schema-specific annotations.
Beyond the cost, even identifying a sufficient number of speakers for data generation and quality control can be difficult.
Most studies have thus focused on collecting data for one such low-resource language and determining the utility of multilingual models or cross-lingual learning from more readily available languages.
Moreover, such datasets are often isolated collections, creating an environment of multiple datasets not easily comparable across the different languages or tasks.
There have been exceptions, such as SQuAD \citep{rajpurkar2016squad} and XQuAd \citep{Artetxe:etal:2019}, ATIS \citep{price-1990-evaluation}, its Hindi and Turkish extension \citep{8461905}, and MultiATIS++ \citep{xu-etal-2020-end}, and Snips \citep{coucke2018snips} with its addition of French \citep{saade2019spoken}, where researchers have extended popular English benchmark datasets to new languages.
This work focuses on the general multi-domain NLU task and builds off the SLURP \citep{bastianelli-etal-2020-slurp} benchmark dataset to extend to an unprecedented 50 new languages.

\begin{table*}[!ht]
\resizebox{\textwidth}{!}{%
\begin{tabular}{llllll}
\toprule
Name                                         & \# Lang & Utt per Lang & Domains & Intents & Slots                                                                                                           \\ \midrule
\M                                           & 51           & 19,521 & 18 & 60 & 55 \\
SLURP \citep{bastianelli-etal-2020-slurp}                                        & 1           & 16,521 & 18 & 60 & 55                                                \\
NLU Evaluation Data \citep{liu2019benchmarking}                          & 1            & 25,716 & 18 & 54 & 56                                         \\
Airline Travel Information System (ATIS) \citep{price-1990-evaluation}    & 1 & 5,871 & 1 & 26 & 129                                                           \\
ATIS with Hindi and Turkish \citep{8461905}                                 & 3 & 1,315-5,871 & 1 & 26 & 129                                                   \\
MultiATIS++ \citep{xu-etal-2020-end}                                 & 9            & 1,422-5,897 & 1 & 21-26 & 99-140                                                    \\
Snips \citep{coucke2018snips}                                       & 1 & 14,484 & - & 7 & 53                                                                           \\
Snips with French \citep{saade2019spoken}                                       & 2            & 4,818 & 2 & 14-15 & 11-12                                                                            \\
Task Oriented Parsing (TOP) \citep{gupta-etal-2018-semantic-parsing}         & 1            & 44,873 & 2 & 25 & 36                  \\
Multilingual Task-Oriented Semantic Parsing (MTOP) \citep{li-etal-2021-mtop} & 6           & 15,195-22,288 & 11 & 104-113 & 72-75                                                \\
Cross-lingual Multilingual Task Oriented Dialog \citep{schuster-etal-2019-cross-lingual}  & 3            & 5,083-43,323 & 3 & 12 & 11                                          \\
Microsoft Dialog Challenge \citep{li2018microsoft}                  & 1            & 38,276 & 3 & 11 & 29 \\
Fluent Speech Commands (FSC) \citep{lugosch2019speech}        & 1            & 30,043 & - & 31 & -                 \\
Chinese Audio-Textual Spoken Language Understanding (CATSLU) \citep{10.1145/3340555.3356098}         & 1            & 16,258 & 4 & - &  94                 \\
\bottomrule
\end{tabular}}
\caption{\label{table:NLUDatasets} Selected NLU benchmark datasets with number of languages, utterances per language, domain count, intent count, and slot count.}
\end{table*}

For the task of NLU, the ATIS dataset has been popular in the NLP community since its first release.
MultiATIS++ was one of the first efforts to extend an NLU dataset across a significant number of languages (nine), yet remained in the limited domain of airline bookings.
While proving an asset, it has been questioned what is left to learn from such a dataset \citep{tur2010left}.
Facebook released a general Intelligent Virtual Assistant (IVA) dataset across the domains of Alarm, Reminder, and Weather \citep{schuster-etal-2019-cross-lingual} created for the purpose of demonstrating cross-lingual transfer learning; and so did not need to be parallel or have an equal number of datapoints, resulting in far fewer examples in Thai (5k) compared to Spanish (7.6k) and English (43k).
The Snips datasets (both the original English only and the English and French releases) are most similar to the NLU contained in the \M~dataset, spanning smart home and music domains for a generic voice-based virtual assistant.

The first iteration for the foundation of the \M~dataset was the NLU Evaluation Benchmarking Dataset, with 25k utterances across 18 domains \citep{liu2019benchmarking}.
The authors updated the dataset and added audio and ASR transcriptions in the release of the Spoken Language Understanding Resource Package (SLURP) \citep{bastianelli-etal-2020-slurp}, allowing for full end-to-end Spoken Language Understanding (SLU) evaluation similar to the Fluent Speech Commands dataset \citep{lugosch2019speech} and Chinese Audio-Textual Spoken Language Understanding (CATSLU) \citep{10.1145/3340555.3356098}.
An overview of selected existing NLU datasets can be seen in Table~\ref{table:NLUDatasets}.

We release the \M~dataset along with baselines from large pre-trained models fine-tuned on the NLU slot and intent prediction tasks.
Early cross-lingual and multilingual NLU modeling approaches used projection or alignment methods \citep{yarowsky-etal-2001-inducing}, focusing on string matching, edit distance, or consonant signatures \citep{ehrmann-etal-2011-building}, lookup lexicons for low-resource languages \citep{mayhew-etal-2017-cheap}, and aligning \citep{xie-etal-2018-neural} or jointly training word embeddings \citep{singla-etal-2018-multi}.
More recently, researchers have borrowed encoders from pre-trained neural translation models before building subsequent classifiers and NER models \citep{eriguchi2018zeroshot, schuster-etal-2019-cross-lingual}, also focusing on language-agnostic and language specific features to learn what information to share between languages \citep{chen-etal-2019-multi-source}.
Generative parsing has been demonstrated using sequence-to-sequence models and pointer networks \citep{dontparsegenerate}.
With the rise of BERT and large pre-trained language models, we have also seen impressive demonstrations of zero-shot performance, where subword tokenization WordPiece overlap helps but is not even necessary to realize improvements \citep{pires-etal-2019-multilingual, k2020crosslingual}, as well as production multilingual NLU improvements with distillation and full fine-tuning \citep{FitzGerald2022AlexaTM}.
The translation task has then been incoporated in the pre-training \citep{wang-etal-2021-exploring-cross} of these models or even as part of the final NLU hypothesis for streamlined multilingual production systems \citep{fitzgerald2020stil}.
Researchers have propped up training data by translating and projecting labels into the target language \citep{xu-etal-2020-end} and discovered more sophisticated approaches to alignment such as translate and fill using mT5 to train the filler \citep{nicosia-etal-2021-translate-fill}.
Recent work has even delved into the application of these techniques to lower-resource languages such as Persian. For example, ParsiNLU explores a variety of NLU tasks for Parsi, fine-tuning mT5 of various sizes \citep{khashabi2021parsinlu}. Similarly these techniques have also been used, even a bit earlier, for text summarization \citep{Farahanipersian2021}.

\section{Language Selection and Linguistic Analysis}
\label{sect:linguistics}

\subsection{Language Selection}

The languages in \M{} were chosen according to the following considerations. First, we acquired cost and worker availability estimates for over 100 languages, providing a constraint to our choices given our fixed budget.
Second, we determined existing languages available in major virtual assistants, such that the dataset could be used to benchmark today's systems.
Third, we categorized the full pool of languages according to their genera as taken from the World Atlas of Linguistic Structures (WALS) database \citep{wals}, where a genus is a language group that is clear to most linguists without systematic comparative analysis.
Genus is a better indicator of typological diversity, which we sought to maximize, than language family \citep{Dryer-1989}.
Fourth, we used the eigenvector centrality of Wikipedia articles, tweets, and book translations \citep{links-that-speak} as proxies for the internet influence and thus the resource availability of a given language, particularly for self-supervised pretraining applications, and we chose languages spanning the breadth of resource availability.
Fifth, we examined the script of each language, seeking to increase script diversity to drive experimentation in tokenization and normalization.

Ultimately, we created 50 new, distinct text corpora, representing 49 different spoken languages.
Mandarin Chinese was collected twice, once with native speakers who use the traditional set of characters, and once with native speakers who use the modern simplified set of characters.
There are 14 language families in the dataset.
The term ``language family'' usually refers to a group of languages which are known to be genetically related, that is, they all descend from a common ancestor language.
In \M{}, we also include ``language isolates'' as families.
These are languages that have no clear relationship to any known language.
Our choices are given in Table \ref{table:Langs}.

\begin{table*}[!ht]
\resizebox{\textwidth}{!}{%
\begin{tabular}{llll|llll|llll}
\toprule
Code & Name & Script & Genus & Code & Name & Script & Genus & Code & Name & Script & Genus \\
\midrule
af-ZA & Afrikaans & Latn & Germanic & hy-AM & Armenian & Armn & Armenian & pl-PL & Polish & Latn & Slavic \\
am-ET & Amharic & Ethi & Semitic & id-ID & Indonesian & Latn & Malayo-Sumbawan & pt-PT & Portuguese & Latn & Romance \\
ar-SA & Arabic & Arab & Semitic & is-IS & Icelandic & Latn & Germanic & ro-RO & Romanian & Latn & Romance \\
az-AZ & Azerbaijani & Latn & Turkic & it-IT & Italian & Latn & Romance & ru-RU & Russian & Cyrl & Slavic \\
bn-BD & Bengali & Beng & Indic & ja-JP & Japanese & Jpan & Japanese & sl-SI & Slovenian & Latn & Slavic \\
cy-GB & Welsh & Latn & Celtic & jv-ID & Javanese & Latn & Javanese & sq-AL & Albanian & Latn & Albanian \\
da-DK & Danish & Latn & Germanic & ka-GE & Georgian & Geor & Kartvelian & sv-SE & Swedish & Latn & Germanic \\
de-DE & German & Latn & Germanic & km-KH & Khmer & Khmr & Khmer & sw-KE & Swahili & Latn & Bantoid \\
el-GR & Greek & Grek & Greek & kn-IN & Kannada & Knda & Southern Dravidian & ta-IN & Tamil & Taml & Southern Dravidian \\
en-US & English & Latn & Germanic & ko-KR & Korean & Kore & Korean & te-IN & Telugu & Telu & South-Central Dravidian \\
es-ES & Spanish & Latn & Romance & lv-LV & Latvian & Latn & Baltic & th-TH & Thai & Thai & Kam-Tai \\
fa-IR & Persian & Arab & Iranian & ml-IN & Malayalam & Mlym & Southern Dravidian & tl-PH & Tagalog & Latn & Greater Central Philippine \\
fi-FI & Finnish & Latn & Finnic & mn-MN & Mongolian & Cyrl & Mongolic & tr-TR & Turkish & Latn & Turkic \\
fr-FR & French & Latn & Romance & ms-MY & Malay & Latn & Malayo-Sumbawan & ur-PK & Urdu & Arab & Indic \\
he-IL & Hebrew & Hebr & Semitic & my-MM & Burmese & Mymr & Burmese-Lolo & vi-VN & Vietnamese & Latn & Viet-Muong \\
hi-IN & Hindi & Deva & Indic & nb-NO & Norwegian & Latn & Germanic & zh-CN & Mandarin & Hans & Chinese \\
hu-HU & Hungarian & Latn & Ugric & nl-NL & Dutch & Latn & Germanic & zh-TW & Mandarin & Hant & Chinese \\
\bottomrule
\end{tabular}
}
\caption{\label{table:Langs} The 51 languages of \M{}, including scripts and genera.}
\end{table*}

\subsection{Scripts} \label{sect:Scripts}

There are 21 distinct scripts used in the dataset.
The majority of languages in \M{} (28 including English) use some variety of the Latin alphabet, which is also the most widely used script in the world.
The Arabic script is used for three languages, the Cyrillic script for two languages, and the remaining 18 languages have ``unique'' scripts, in the sense that only one language in the dataset uses that script.
Fourteen scripts are unique to a single language, although they may belong to a larger family of writing systems.
For example, the Dravidian languages in \M{} have their own scripts, but are all members of the general Brahmi class of scripts.
The other two scripts are unique in that only one language in the dataset uses them, but they are more widely used in the real world: Ge'ez and Chinese.
Ge'ez is represented by Amharic in the dataset, but is used for several languages in East Africa, such as Tigrinya.
The Chinese script is represented by Mandarin, but is used by other languages in China such as Cantonese.

\subsection{Sentence Types} \label{sect:SentenceTypes}

\M{} consists of utterances directed at a device, rather than a person, which has some consequences for the type of linguistic patterns it contains.
Specifically, the corpus primarily consists of interrogatives (i.e., questions) and imperatives (commands or requests).
There are relatively few declarative utterances in the set.
This is in contrast to many large datasets from other sources (e.g., wikipedia, movie scripts, newspapers) which contain a high proportion of declaratives, since the language is collected from situations where humans are communicating with humans.

In the context of a voice assistant, a user typically asks a device to perform an action or answer a question, so declaratives are less common.
For instance, a person might use an imperative ``tell me if it calls for rain today'' or ask a question ``will it rain today,'' but they would not tell their device ``it's raining today.''
When declaratives are used with voice assistants, they generally have the pragmatic effect of a directive. For instance, a virtual assistant can respond to the declarative ``it's cold in here'' by turning up the temperature \cite{thattai_tur_natarajan_2020}.
Although syntactically it looks like a declarative, such an utterance has the force of an imperative.

The standard unit of analysis in linguistics is the declarative sentence, and there is relatively less known about imperatives and questions.
\M{} presents an opportunity to study these sentence forms, and the parallel nature of the corpus makes cross-linguistic comparisons even easier.

\subsection{Word Order} \label{sect:WordOrder}

Languages have intricate rules for ordering words depending on the word-type and  sentence-type.
In English, the word order for statements (``you are leaving'') is different from questions (``are you leaving?'').
This is not mandatory, and sometimes the pitch of the voice is enough to indicate a question (e.g. ``you're leaving?'' with a rising intonation).

When considering word order at a typological level, it is common to simplify the situation and consider only affirmative declarative sentences and only three grammatical elements: the verb (V), its subject (S), and its object (O).
This makes for six possible word orders: SVO, SOV, VOS, VSO, OVS, and OSV.
All six orders have been documented, although the overwhelming majority of languages use Subject-initial ordering, while Object-initial ordering is extremely rare.

In \M{}, 39 languages are subject-initial (24 SVO and 15 SOV), while only three are verb-initial (VSO specifically).
No object-initial languages are represented.
Five languages are marked in WALS as having no preferred word order, and four do not have any word order data at all.

\subsection{Imperative Marking} \label{sect:ImperativeMarking}

The languages in \M{} have a variety of ways of indicating the imperative mood of an utterance.
The majority of them (33) use some kind of verb morphology, such as adding a suffix.
About half of those languages (18) have distinct imperative marking for singular or plural addressees.
The utterances in \M{} are technically directed at a single addressee, the voice assistant, but since some languages use the plural as an indicator of politeness (see below) all varieties of imperatives will likely occur in this dataset.
There are ten languages without any special morphology, and they indicate imperative through other means, such as word order or vocabulary choice.

Ten languages in the dataset have a specialized distinction between imperatives, for commands directed at another individual, and ``hortatives'', where the command also includes the speaker.
English verbs are not directly marked for hortative, but the auxiliary verb ``let'' can convey the mood instead.
For example, ``write this down'' is an imperative and only the addressee need write anything, while ``let's write this down'' is a hortative and the speaker is also expected to write.
The pervasiveness of hortatives in the context of a voice assistant is an open question.

Four languages have ``optative'' moods, which are subtly different from imperatives.
In the optative, a speaker expresses a wish or desire, as opposed to giving a direct command.
However, in the right context, an optative may carry the same pragmatic weight as an imperative, and strongly imply that someone ought to do something.
English has no specific optative form, but a similar mood can be conveyed using conditionals.
For example, ``buy this bag for me'' is an imperative while ``if only someone would buy me this bag'' is closer to an optative.
Optative forms are not well studied in linguistics, as they require specific contexts which can be difficult to create during field work, but they may be more common in device-directed utterances.

Lastly, some languages distinguish between imperatives, when telling someone to do something, and ``prohibitives'', when telling someone not to do something.
In the \M{} set, there are 18 languages with specialized negative particles which can only co-occur with imperative verbs.
Vietnamese for instance uses the words ``chăng''  or ``không'' to negate declarative sentences, but uses ``chó'' or ``dung'' to negate imperatives.
Another ten languages have special verbs for the prohibitive, although these may overlap with other grammatical features of the language. In Spanish, for example, the prohibitive form of a verb is the same as the subjunctive form. 

\subsection{Politeness} \label{sect:Politeness}

Many languages encode different levels of politeness through their use of pronouns.
Many European languages distinguish between ``familiar'' and ``formal'' pronouns, with the ``formal'' pronouns often morphologically identical to a plural.
In French, the second-person singular ``tu'' is used between friends, while the second-person plural ``vous'' is used when speaking to a group, or to an individual of higher social rank (such as an employee to a manager).
These politeness systems are heavily influenced by social context, and the \M{} dataset gives us a chance to see how people adapt their language when speaking to a virtual assistant instead of another human. 

Nearly half of the languages in \M{} (21) make a two-way formal/informal distinction in their second-person pronouns.
This is probably due to the fact that most \M{} languages are European, and the binary politeness distinctions are the most common strategy in that family.
A further eight languages have more than two levels of formality, such as informal, formal, and honorific.
Seven languages have an ``avoidance'' strategy, which means that pronouns are omitted entirely in a polite scenario.
Finally, eleven languages have no data on politeness in WALS at all.

\section{Collection Setup and Execution}

\subsection{Heldout Evaluation Split}
We randomly sampled a subset of the English seed data which was then paraphrased by professional annotators, resulting in new, more challenging utterances, including 49\% more slots per utterance. 
\ifreview
\else
These utterances were localized along with the other splits to be used as a held out evaluation set for the Massively Multilingual NLU-22 competition and workshop \footnote{\href{http://mmnlu-22.github.io}{mmnlu-22.github.io}}.
\fi

\subsection{Vendor Selection and Onboarding}

The \M~dataset was collected using a customized workflow powered by Amazon MTurk.
We required a vendor pool with the capability and resources to collect a large multilingual dataset.
Our original vendor pool consisted of five vendors adjudicated based on previous engagements.
This vendor pool was reduced to three based on engagement and resource availability.
Vendors for each language were selected based on their resource availability and proposed cost.
A majority of languages were supported by a single vendor, while some languages required cross-vendor support to be completed with the required quality and within the required timeline.
 
We offered two mechanisms to vendors for evaluating workers to be selected for each language.
The first, which was used to select workers for the translation task, was an Amazon  MTurk-hosted fluency test where workers listen to questions and statements in the relevant language and were evaluated using a multiple-choice questionnaire.
The second, which was used to select workers for the judgment task, was a test with a set of three judgments that the vendor could use to assess if workers were able to detect issues in the translated utterances.
In order to further improve worker selection quality, we created a translator quiz using the Amazon MTurk instructions that were created for translation and judgment tasks, coupled with customized local-language examples.
The workers were required to prove that they understood the instructions for the project based on a series of questions. 
 
Before commencing operations, an initial pilot run of this customized workflow was completed in three languages.
A few workers per vendor were chosen to engage in this exercise.
The pilot run helped improve clarity of instructions, determine reporting methods, and share open questions.

\subsection{Collection Workflows} \label{sect:localization-workflow}

The collection was conducted by locale on an individual utterance level.
Each utterance from the ``train,'' ``dev,'' ``test,'' and ``heldout'' splits of the SLURP dataset went through two sequential task workflows and a judgment workflow.
The first task is slot translation or localization (see Figure~\ref{fig:slot}).
Workers are presented the entire utterance with colored highlighting of the slot values for the utterance (if any) and then presented with each slot value and its corresponding label individually.
The worker is asked to either localize or translate the slot, depending on whether the value should be translated (e.g., ``tomorrow'') or localized (e.g., the movie ``La La Land'', which in French is ``Pour l'amour d'Hollywood.'' Other entities like regionally known songs or artists could also be localized to a more relevant, known song or artist for that language or region).
There is also an option to keep the slot as is, such as for names (e.g., ``Taylor Swift'') or proper nouns where the original English spelling should be retained.
The metadata of the released dataset includes whether the worker elected to ``localize,'' ``translate,'' or keep the slot ``unchanged,'' primarily for the purposes of researchers evaluating machine translation systems, where it would be unreasonable to expect the system to ``localize'' to a specific song name the worker selected.

After the slot task, the second worker is asked to translate or localize the entire phrase using the slot task output provided by the first worker (see Figure~\ref{fig:phrase}).
The phrase worker can decide to keep the slot as it was translated, modify it, or remove it entirely if it is not relevant for the language in that scenario.
This worker is also responsible for aligning grammatical genders or prepositional affixes to any of the slots.

Note that this two-step system alleviates the annotation burden often encountered with such work.
Traditionally in such collections, workers would be given a light annotation guide and asked to highlight spans of the slots in a translated or localized utterance.
In this system, the first step of slot translation and subsequent insertion obviates the need for workers to understand nuanced span notation, which can be complex for highly inflected languages (prepositions outside the span in English would not be carried over in the localization, but would be in the traditional span annotation workflow).

\subsection{Quality Assurance} \label{sect:QA}

The output of the second workflow (the fully localized utterance) is judged by three workers for (1) whether the utterance matches the intent semantically, (2) whether the slots match their labels semantically, (3) grammaticality and naturalness, (4) spelling, and (5) language identification---English or mixed utterances are acceptable if that is natural for the language, but localizations without any tokens in the target language were not accepted.
See Figure~\ref{fig:judgment} for how this is presented to the Amazon MTurk worker.
These judgments are also included in the metadata of the dataset.
In addition to the workers judging each other's work, the collection system had alarms in place for workers with high rejection rates, high rates of slot deletion, and high rates of English tokens in the translations.
Workers were also monitored to see if their tasks were primarily machine translated.
Such workers were removed from the pool and all of their work was resubmitted to be completed by the other workers.
Additionally, the authors performed several deep dives into languages with which they were familiar.

\section{Model Benchmarking}
\label{sect:modeling}

\subsection{Setup}\label{sect:setup}

As initial model benchmarks, we fine-tuned publicly-available pre-trained language models on the \M~dataset and evaluated them on intent classification and slot filling.
Our models of choice for this exercise were XLM-Roberta (XLM-R; \citealt{conneau-etal-2020-unsupervised}) and mT5 \citep{xue-etal-2021-mt5}.

In the case of XLM-R, we utilized the pre-trained encoder with two separate classification heads trained from scratch, based on JointBERT \citep{Chen2019BERTFJ}. The first classification head used the pooled output from the encoder to predict the intent and the second used the sequence output to predict the slots. As pooling for the intent classification head, we experimented with using hidden states from the first position, averaged hidden states across the sequence, and the maximally large hidden state from the sequence.

\begin{table*}[]
\subfloat[][Test results when using the full training set]{
\resizebox{\linewidth}{!}{%
\begin{tabular}{c|ccc|ccc|ccc}
\toprule
Model & \multicolumn{3}{c}{Intent Acc (\%)} &  \multicolumn{3}{c}{Slot F1 (\%)} &  \multicolumn{3}{c}{Exact Match Acc (\%)} \\
 & High & Low & Avg & High & Low & Avg & High & Low & Avg \\
\midrule
mT5 Base & 87.9 $\pm$ 1.2 & 79.0 $\pm$ 1.5 & 85.3 $\pm$ 0.2 & 86.8 $\pm$ 0.7 & 67.6 $\pm$ 0.4 & 76.8 $\pm$ 0.1 & 73.4 $\pm$ 1.6 & 58.3 $\pm$ 1.8 & 66.6 $\pm$ 0.2 \\
Text-to-Text & en-US & km-KH & & th-TH & ja-JP & & th-TH & ja-JP & \\
mT5 Base & 89.0 $\pm$ 1.1 & 79.1 $\pm$ 1.5 & 86.1 $\pm$ 0.2 & 85.7 $\pm$ 0.7 & 64.5 $\pm$ 0.4 & 75.4 $\pm$ 0.1 & 72.3 $\pm$ 1.6 & 57.8 $\pm$ 1.8 & 65.9 $\pm$ 0.2 \\
Encoder-Only & en-US & km-KH & & th-TH & ja-JP & & th-TH & ja-JP & \\
XLM-R Base & 88.3 $\pm$ 1.2 & 77.2 $\pm$ 1.5 & 85.1 $\pm$ 0.2 & 83.5 $\pm$ 0.7 & 63.3 $\pm$ 0.4 & 73.6 $\pm$ 0.1 & 70.1 $\pm$ 1.6 & 55.8 $\pm$ 1.8 & 63.7 $\pm$ 0.2 \\
 & en-US & km-KH & & th-TH & ja-JP & & th-TH & ja-JP & \\

\bottomrule
\end{tabular}
}
}
\newline
\subfloat[][Zero-shot test results after training only on en-US]{
\resizebox{\linewidth}{!}{%
\begin{tabular}{c|ccc|ccc|cccl}
\toprule
Model & \multicolumn{3}{c}{Intent Acc (\%)} &  \multicolumn{3}{c}{Slot F1 (\%)} &  \multicolumn{3}{c}{Exact Match Acc (\%)} \\
 & High & Low & Avg & High & Low & Avg & High & Low & Avg \\
\midrule
mT5 Base & 79.9 $\pm$ 1.4 & 25.7 $\pm$ 1.6 & 62.9 $\pm$ 0.2 & 64.3 $\pm$ 0.7 & 13.9 $\pm$ 0.3 & 44.8 $\pm$ 0.1 & 53.2 $\pm$ 1.8 & 9.4 $\pm$ 1.0 & 34.7 $\pm$ 0.2 \\
Text-to-Text & nl-NL & ja-JP & & de-DE & ja-JP & & sv-SE & ja-JP & \\
mT5 Base & 76.4 $\pm$ 1.5 & 27.1 $\pm$ 1.6 & 61.2 $\pm$ 0.2 & 59.5 $\pm$ 1.0 & 6.3 $\pm$ 0.2 & 41.6 $\pm$ 0.1 & 44.3 $\pm$ 1.8 & 4.2 $\pm$ 0.7 & 28.8 $\pm$ 0.2 \\
Encoder-Only & nl-NL & ja-JP & & th-TH & ja-JP & & sv-SE & ja-JP & \\
XLM-R Base & 85.2 $\pm$ 1.3 & 44.8 $\pm$ 1.8 & 70.6 $\pm$ 0.2 & 68.4 $\pm$ 0.7 & 15.4 $\pm$ 0.3 & 50.3 $\pm$ 0.1 & 57.9 $\pm$ 1.8 & 9.8 $\pm$ 1.1 & 38.7 $\pm$ 0.2 \\
 & sv-SE & ja-JP & & sv-SE & ja-JP & & sv-SE & ja-JP & \\
 
\bottomrule
\end{tabular}
}
}
\caption{Modeling results for (a) training runs on the full training dataset and (b) zero-shot training runs, in which training was performed only with en-US data, validation was performed with all locales, and testing was performed on all locales except for en-US. Each table includes the highest locale, the lowest locale, and locale-averaged results for intent accuracy, micro-averaged slot F1 score, and exact match accuracy. Intervals for 95\% confidence are given assuming normal distributions.}
\label{tab:test_results}
\end{table*}

With mT5, we explored two separate architectures.
In one architecture, we only used the pre-trained encoder extracted from mT5, and we trained two classification heads from scratch similarly to the XLM-R setup.
We refer to this setup as mT5 Encoder-Only.
In the other architecture, we used the full sequence-to-sequence mT5 model in text-to-text mode, where the input is ``\texttt{Annotate:}'' followed by the unlabeled utterance.
The decoder output is a sequence of labels (including the \texttt{Other} label) for all of the tokens followed by the intent.
We did not add the slots and intents to the vocabulary, but we instead allowed them to be tokenized into subwords.
We refer to this model as mT5 Text-to-Text.
For all models, we used the Base size, which corresponds to 270M parameters for XLM-R, 258M parameters for mT5 Encoder-Only, and 580M parameters for mT5 Text-to-Text, including 192M parameters for embeddings for all three.

For each model, we performed 128 trials of hyperparameter tuning using the Tree of Parzen Estimators algorithm and Asynchronous Successive Halving Algorithm (ASHA) \cite{Li2018MassivelyPH} for scheduling, which are both part of the \texttt{hyperopt} library \citep{pmlr-v28-bergstra13} integrated into the \texttt{ray[tune]} library \cite{liaw2018tune}, which is itself integrated into the \texttt{Trainer} from the \texttt{transformers} library \citep{wolf-etal-2020-transformers}, which we used for modeling and for our pretrained models.
Our hyperparameter search spaces, sampling types, and final choices are given in Table~\ref{tab:model_hyperparams}.
We trained our models with the Adam optimizer \citep{kingma2017adam} and chose the best performing model checkpoint based on overall exact match accuracy across all locales. 
Hyperparameter tuning and fine-tuning was performed using single p3dn.24xlarge instances (8 x Nvidia v100) for XLM-R and mT5 Text-to-Text and a single g4dn.metal instance (8 x Nvidia T4) for mT5 Encoder-Only. Hyperparameter tuning times were less than 4 days per model and training times were less than 1 day per model.

Our dataset includes several languages where white spacing is not used as a word delimiter. In some cases, spaces do occur, but they might serve as phrase delimiters or denote the end of a sentence.
Three of these written languages, Japanese, Chinese (Traditional), and Chinese (Simplified), do not use spaces anywhere except to identify the end of a sentence.
For these languages, we separate each character in the unlabeled input with a whitespace.
We leave exploration of more sophisticated techniques (such as MeCab for Japanese; \citealt{Kudo2005MeCabY}) to future work.
We use the default spacing provided by annotators for all other languages.

Zero-shot performance was also assessed, in which the models were trained on English data, validation was performed on all languages, and testing was performed on all non-English locales.

\subsection{Results and Analysis}

Table~\ref{tab:test_results} shows the results for each model and training setup, including those for the best performing locale, the worst performing locale, and locale-averaged results for intent accuracy, micro-averaged slot F1 score, and exact match accuracy.
Zero-shot exact match performance is 25-37 points worse than that of full-dataset training runs.
Additionally, the variance in task performance across locales is significantly greater for the zero-shot setup than for full-dataset training.
For example, there is a 15 point difference in exact match accuracy between the highest and lowest locales for mT5 Text-to-Text when using the full training set, while the gap expands to 44 points with zero-shot.

We compared the pretraining data quantities by language for XLM-R to its per-language task performance values, and in the zero shot setup, we found a Pearson correlation of 0.54 for exact match accuracy, 0.58 for intent accuracy, and 0.46 for micro-averaged slot F1 score. In the full dataset training setup, the correlations decrease to 0.42 for exact match accuracy, 0.47 for intent accuracy, and 0.24 for micro-averaged slot F1 score. This suggests that the constant per-language data quantities in \M{} help to mitigate the effects of the language-skewed pretraining data distribution.

In Thai, for which spacing is optional, the model can learn from artificial spacing in the input (around where the slots will be) to improve task performance. For Khmer, the workers had a difficult time adapting their translations and localizations to properly-slotted outputs given the space-optional nature of the language. Additionally, for Japanese and Chinese, we added spaces between all characters when modeling. These single-character inputs differ from the non-spaced inputs used during pretraining, which would be chunked into groups of characters by the tokenizer with corresponding embeddings. By splitting into single characters, we don't allow the model to the use embeddings learned for chunks of characters. This is a likely major cause of the drop in exact match accuracy for Japanese from 58.3\% when training on the full dataset to 9.4\% for zero shot. In the zero shot setup, the model relies solely on pretrained data representations, and individually-spaced characters are rare in the pretraining data. That said, character spacing was necessary in order to properly assign the slots to the right characters. As mentioned in Section \ref{sect:setup}, we leave exploration of more sophisticated spacing techniques for slot filling (such as MeCab; \citealt{Kudo2005MeCabY}) to future work.

Discounting for artificial spacing effects, Germanic genera and Latin scripts performed the best overall (See Appendix \ref{sect:perf_lang_char}), which is unsurprising given the amount of pretraining data for those genera and scripts, as well as the quantity of Germanic and Latin-script languages in \M{}. Within the Germanic genera, Swedish, English, Danish, Norwegian, and Dutch all performed comparably (within 95\% confidence bounds) for exact match accuracy. Icelandic was the lowest-performing Germanic language, likely due to a lack of pretraining data, as well as to its linguistic evolution away from the others due to isolated conditions.

\section{Conclusion}

We have released a truly \M~ multilingual dataset for NLU spanning 51 typologically diverse languages. Our hope is that \M~ will encourage many new innovations in massively multilingual NLU, other NLP tasks such as machine translation, and new linguistic analyses, such as with imperative morphologies. 

%

\bibliography{anthology,centurion-emnlp-22}

\begin{thebibliography}{64}
\expandafter\ifx\csname natexlab\endcsname\relax\def\natexlab#1{#1}\fi

\bibitem[{Artetxe et~al.(2019)Artetxe, Ruder, and Yogatama}]{Artetxe:etal:2019}
Mikel Artetxe, Sebastian Ruder, and Dani Yogatama. 2019.
\newblock \href {http://arxiv.org/abs/1910.11856} {On the cross-lingual
  transferability of monolingual representations}.
\newblock \emph{CoRR}, abs/1910.11856.

\bibitem[{Bastianelli et~al.(2020)Bastianelli, Vanzo, Swietojanski, and
  Rieser}]{bastianelli-etal-2020-slurp}
Emanuele Bastianelli, Andrea Vanzo, Pawel Swietojanski, and Verena Rieser.
  2020.
\newblock \href {https://doi.org/10.18653/v1/2020.emnlp-main.588} {{SLURP}: A
  spoken language understanding resource package}.
\newblock In \emph{Proceedings of the 2020 Conference on Empirical Methods in
  Natural Language Processing (EMNLP)}, pages 7252--7262, Online. Association
  for Computational Linguistics.

\bibitem[{Bergstra et~al.(2013)Bergstra, Yamins, and Cox}]{pmlr-v28-bergstra13}
James Bergstra, Daniel Yamins, and David Cox. 2013.
\newblock \href {https://proceedings.mlr.press/v28/bergstra13.html} {Making a
  science of model search: Hyperparameter optimization in hundreds of
  dimensions for vision architectures}.
\newblock In \emph{Proceedings of the 30th International Conference on Machine
  Learning}, volume~28 of \emph{Proceedings of Machine Learning Research},
  pages 115--123, Atlanta, Georgia, USA. PMLR.

\bibitem[{Chen et~al.(2019{\natexlab{a}})Chen, Zhuo, and Wang}]{Chen2019BERTFJ}
Qian Chen, Zhu Zhuo, and Wen Wang. 2019{\natexlab{a}}.
\newblock Bert for joint intent classification and slot filling.
\newblock \emph{ArXiv}, abs/1902.10909.

\bibitem[{Chen et~al.(2019{\natexlab{b}})Chen, Awadallah, Hassan, Wang, and
  Cardie}]{chen-etal-2019-multi-source}
Xilun Chen, Ahmed~Hassan Awadallah, Hany Hassan, Wei Wang, and Claire Cardie.
  2019{\natexlab{b}}.
\newblock \href {https://doi.org/10.18653/v1/P19-1299} {Multi-source
  cross-lingual model transfer: Learning what to share}.
\newblock In \emph{Proceedings of the 57th Annual Meeting of the Association
  for Computational Linguistics}, pages 3098--3112, Florence, Italy.
  Association for Computational Linguistics.

\bibitem[{Conneau et~al.(2020)Conneau, Khandelwal, Goyal, Chaudhary, Wenzek,
  Guzm{\'a}n, Grave, Ott, Zettlemoyer, and
  Stoyanov}]{conneau-etal-2020-unsupervised}
Alexis Conneau, Kartikay Khandelwal, Naman Goyal, Vishrav Chaudhary, Guillaume
  Wenzek, Francisco Guzm{\'a}n, Edouard Grave, Myle Ott, Luke Zettlemoyer, and
  Veselin Stoyanov. 2020.
\newblock \href {https://doi.org/10.18653/v1/2020.acl-main.747} {Unsupervised
  cross-lingual representation learning at scale}.
\newblock In \emph{Proceedings of the 58th Annual Meeting of the Association
  for Computational Linguistics}, pages 8440--8451, Online. Association for
  Computational Linguistics.

\bibitem[{Coucke et~al.(2018)Coucke, Saade, Ball, Bluche, Caulier, Leroy,
  Doumouro, Gisselbrecht, Caltagirone, Lavril, Primet, and
  Dureau}]{coucke2018snips}
Alice Coucke, Alaa Saade, Adrien Ball, Th{\'e}odore Bluche, Alexandre Caulier,
  David Leroy, Cl{\'e}ment Doumouro, Thibault Gisselbrecht, Francesco
  Caltagirone, Thibaut Lavril, Ma{\"e}l Primet, and Joseph Dureau. 2018.
\newblock \href {http://arxiv.org/abs/1805.10190} {Snips voice platform: an
  embedded spoken language understanding system for private-by-design voice
  interfaces}.

\bibitem[{Cruz and Cheng(2020)}]{cruz2020establishing}
Jan Christian~Blaise Cruz and Charibeth Cheng. 2020.
\newblock \href {http://arxiv.org/abs/2005.02068} {Establishing baselines for
  text classification in low-resource languages}.

\bibitem[{Devlin(2018)}]{devlinmBERT}
Jacob Devlin. 2018.
\newblock \href
  {https://github.com/google-research/bert/blob/master/multilingual.md}
  {Multiligual bert}.

\bibitem[{Dryer(1989)}]{Dryer-1989}
Matthew~S. Dryer. 1989.
\newblock Large linguistic areas and language sampling.
\newblock \emph{Studies in Language}, 13:257--292.

\bibitem[{Dryer and Haspelmath(2013)}]{wals}
Matthew~S. Dryer and Martin Haspelmath, editors. 2013.
\newblock \href {https://wals.info/} {\emph{WALS Online}}.
\newblock Max Planck Institute for Evolutionary Anthropology, Leipzig.

\bibitem[{Ehrmann et~al.(2011)Ehrmann, Turchi, and
  Steinberger}]{ehrmann-etal-2011-building}
Maud Ehrmann, Marco Turchi, and Ralf Steinberger. 2011.
\newblock \href {https://aclanthology.org/R11-1017} {Building a multilingual
  named entity-annotated corpus using annotation projection}.
\newblock In \emph{Proceedings of the International Conference Recent Advances
  in Natural Language Processing 2011}, pages 118--124, Hissar, Bulgaria.
  Association for Computational Linguistics.

\bibitem[{Eriguchi et~al.(2018)Eriguchi, Johnson, Firat, Kazawa, and
  Macherey}]{eriguchi2018zeroshot}
Akiko Eriguchi, Melvin Johnson, Orhan Firat, Hideto Kazawa, and Wolfgang
  Macherey. 2018.
\newblock \href {http://arxiv.org/abs/1809.04686} {Zero-shot cross-lingual
  classification using multilingual neural machine translation}.

\bibitem[{Farahani et~al.(2021)Farahani, Gharachorloo, and
  Manthouri}]{Farahanipersian2021}
Mehrdad Farahani, Mohammad Gharachorloo, and Mohammad Manthouri. 2021.
\newblock \href {https://doi.org/10.1109/csicc52343.2021.9420563} {Leveraging
  parsbert and pretrained mt5 for persian abstractive text summarization}.
\newblock \emph{2021 26th International Computer Conference, Computer Society
  of Iran (CSICC)}.

\bibitem[{FitzGerald et~al.(2022)FitzGerald, Ananthakrishnan, Arkoudas,
  Bernardi, Bhagia, Bovi, Cao, Chada, Chauhan, Chen, Dwarakanath, Dwivedi,
  Gojayev, Gopalakrishnan, Gueudre, Hakkani-Tur, Hamza, Hueser, Jose, Khan,
  Liu, Lu, Manzotti, Natarajan, Owczarzak, Oz, Palumbo, Peris, Prakash, Rawls,
  Rosenbaum, Shenoy, Soltan, Sridhar, Tan, Triefenbach, Wei, Yu, Zheng, Tur,
  and Natarajan}]{FitzGerald2022AlexaTM}
Jack FitzGerald, Shankar Ananthakrishnan, Konstantine Arkoudas, Davide
  Bernardi, Abhishek Bhagia, Claudio~Delli Bovi, Jin Cao, Rakesh Chada, Amit
  Chauhan, Luoxin Chen, Anurag Dwarakanath, Satyam Dwivedi, Turan Gojayev,
  Karthik Gopalakrishnan, Thomas Gueudre, Dilek Hakkani-Tur, Wael Hamza,
  Jonathan Hueser, Kevin~Martin Jose, Haidar Khan, Beiye Liu, Jianhua Lu,
  Alessandro Manzotti, Pradeep Natarajan, Karolina Owczarzak, Gokmen Oz, Enrico
  Palumbo, Charith Peris, Chandana~Satya Prakash, Stephen Rawls, Andy
  Rosenbaum, Anjali Shenoy, Saleh Soltan, Mukund~Harakere Sridhar, Liz Tan,
  Fabian Triefenbach, Pan Wei, Haiyang Yu, Shuai Zheng, Gokhan Tur, and Prem
  Natarajan. 2022.
\newblock Alexa teacher model: Pretraining and distilling
  multi-billion-parameter encoders for natural language understanding systems).
\newblock In \emph{Proceedings of the 28th ACM SIGKDD Conference on Knowledge
  Discovery and Data Mining}, KDD. ACM.

\bibitem[{FitzGerald(2020)}]{fitzgerald2020stil}
Jack G.~M. FitzGerald. 2020.
\newblock \href {http://arxiv.org/abs/2010.00760} {Stil -- simultaneous slot
  filling, translation, intent classification, and language identification:
  Initial results using mbart on multiatis++}.

\bibitem[{Goyal et~al.(2021)Goyal, Gao, Chaudhary, Chen, Wenzek, Ju, Krishnan,
  Ranzato, Guzman, and Fan}]{goyal2021flores101}
Naman Goyal, Cynthia Gao, Vishrav Chaudhary, Peng-Jen Chen, Guillaume Wenzek,
  Da~Ju, Sanjana Krishnan, Marc'Aurelio Ranzato, Francisco Guzman, and Angela
  Fan. 2021.
\newblock \href {http://arxiv.org/abs/2106.03193} {The flores-101 evaluation
  benchmark for low-resource and multilingual machine translation}.

\bibitem[{Gupta et~al.(2018)Gupta, Shah, Mohit, Kumar, and
  Lewis}]{gupta-etal-2018-semantic-parsing}
Sonal Gupta, Rushin Shah, Mrinal Mohit, Anuj Kumar, and Mike Lewis. 2018.
\newblock \href {https://doi.org/10.18653/v1/D18-1300} {Semantic parsing for
  task oriented dialog using hierarchical representations}.
\newblock In \emph{Proceedings of the 2018 Conference on Empirical Methods in
  Natural Language Processing}, pages 2787--2792, Brussels, Belgium.
  Association for Computational Linguistics.

\bibitem[{Hu et~al.(2020)Hu, Ruder, Siddhant, Neubig, Firat, and
  Johnson}]{hu2020xtreme}
Junjie Hu, Sebastian Ruder, Aditya Siddhant, Graham Neubig, Orhan Firat, and
  Melvin Johnson. 2020.
\newblock \href {http://arxiv.org/abs/2003.11080} {Xtreme: A massively
  multilingual multi-task benchmark for evaluating cross-lingual
  generalization}.

\bibitem[{K et~al.(2020)K, Wang, Mayhew, and Roth}]{k2020crosslingual}
Karthikeyan K, Zihan Wang, Stephen Mayhew, and Dan Roth. 2020.
\newblock \href {http://arxiv.org/abs/1912.07840} {Cross-lingual ability of
  multilingual bert: An empirical study}.

\bibitem[{Khashabi et~al.(2021)Khashabi, Cohan, Shakeri, Hosseini, Pezeshkpour,
  Alikhani, Aminnaseri, Bitaab, Brahman, Ghazarian, Gheini, Kabiri, Mahabadi,
  Memarrast, Mosallanezhad, Noury, Raji, Rasooli, Sadeghi, Azer, Samghabadi,
  Shafaei, Sheybani, Tazarv, and Yaghoobzadeh}]{khashabi2021parsinlu}
Daniel Khashabi, Arman Cohan, Siamak Shakeri, Pedram Hosseini, Pouya
  Pezeshkpour, Malihe Alikhani, Moin Aminnaseri, Marzieh Bitaab, Faeze Brahman,
  Sarik Ghazarian, Mozhdeh Gheini, Arman Kabiri, Rabeeh~Karimi Mahabadi, Omid
  Memarrast, Ahmadreza Mosallanezhad, Erfan Noury, Shahab Raji, Mohammad~Sadegh
  Rasooli, Sepideh Sadeghi, Erfan~Sadeqi Azer, Niloofar~Safi Samghabadi, Mahsa
  Shafaei, Saber Sheybani, Ali Tazarv, and Yadollah Yaghoobzadeh. 2021.
\newblock \href {http://arxiv.org/abs/2012.06154} {Parsinlu: A suite of
  language understanding challenges for persian}.

\bibitem[{Kingma and Ba(2017)}]{kingma2017adam}
Diederik~P. Kingma and Jimmy Ba. 2017.
\newblock \href {http://arxiv.org/abs/1412.6980} {Adam: A method for stochastic
  optimization}.

\bibitem[{Kudo(2005)}]{Kudo2005MeCabY}
Takumitsu Kudo. 2005.
\newblock Mecab : Yet another part-of-speech and morphological analyzer.

\bibitem[{Lakew et~al.(2020)Lakew, Negri, and Turchi}]{lakew2020low}
Surafel~M. Lakew, Matteo Negri, and Marco Turchi. 2020.
\newblock \href {http://arxiv.org/abs/2003.14402} {Low resource neural machine
  translation: A benchmark for five african languages}.

\bibitem[{Lample and Conneau(2019)}]{lample2019crosslingual}
Guillaume Lample and Alexis Conneau. 2019.
\newblock \href {http://arxiv.org/abs/1901.07291} {Cross-lingual language model
  pretraining}.

\bibitem[{Lewis et~al.(2020)Lewis, Ghazvininejad, Ghosh, Aghajanyan, Wang, and
  Zettlemoyer}]{lewis2020pretraining}
Mike Lewis, Marjan Ghazvininejad, Gargi Ghosh, Armen Aghajanyan, Sida Wang, and
  Luke Zettlemoyer. 2020.
\newblock \href {http://arxiv.org/abs/2006.15020} {Pre-training via
  paraphrasing}.

\bibitem[{Li et~al.(2021)Li, Arora, Chen, Gupta, Gupta, and
  Mehdad}]{li-etal-2021-mtop}
Haoran Li, Abhinav Arora, Shuohui Chen, Anchit Gupta, Sonal Gupta, and Yashar
  Mehdad. 2021.
\newblock \href {https://doi.org/10.18653/v1/2021.eacl-main.257} {{MTOP}: A
  comprehensive multilingual task-oriented semantic parsing benchmark}.
\newblock In \emph{Proceedings of the 16th Conference of the European Chapter
  of the Association for Computational Linguistics: Main Volume}, pages
  2950--2962, Online. Association for Computational Linguistics.

\bibitem[{Li et~al.(2018{\natexlab{a}})Li, Jamieson, Rostamizadeh, Gonina,
  Hardt, Recht, and Talwalkar}]{Li2018MassivelyPH}
Liam Li, Kevin~G. Jamieson, Afshin Rostamizadeh, Ekaterina Gonina, Moritz
  Hardt, Benjamin Recht, and Ameet~S. Talwalkar. 2018{\natexlab{a}}.
\newblock Massively parallel hyperparameter tuning.
\newblock \emph{ArXiv}, abs/1810.05934.

\bibitem[{Li et~al.(2018{\natexlab{b}})Li, Wang, Sun, Panda, Liu, and
  Gao}]{li2018microsoft}
Xiujun Li, Yu~Wang, Siqi Sun, Sarah Panda, Jingjing Liu, and Jianfeng Gao.
  2018{\natexlab{b}}.
\newblock \href {http://arxiv.org/abs/1807.11125} {Microsoft dialogue
  challenge: Building end-to-end task-completion dialogue systems}.

\bibitem[{Liaw et~al.(2018)Liaw, Liang, Nishihara, Moritz, Gonzalez, and
  Stoica}]{liaw2018tune}
Richard Liaw, Eric Liang, Robert Nishihara, Philipp Moritz, Joseph~E Gonzalez,
  and Ion Stoica. 2018.
\newblock Tune: A research platform for distributed model selection and
  training.
\newblock \emph{arXiv preprint arXiv:1807.05118}.

\bibitem[{Liu et~al.(2019{\natexlab{a}})Liu, Eshghi, Swietojanski, and
  Rieser}]{liu2019benchmarking}
Xingkun Liu, Arash Eshghi, Pawel Swietojanski, and Verena Rieser.
  2019{\natexlab{a}}.
\newblock \href {http://arxiv.org/abs/1903.05566} {Benchmarking natural
  language understanding services for building conversational agents}.

\bibitem[{Liu et~al.(2020)Liu, Gu, Goyal, Li, Edunov, Ghazvininejad, Lewis, and
  Zettlemoyer}]{liu2020multilingual}
Yinhan Liu, Jiatao Gu, Naman Goyal, Xian Li, Sergey Edunov, Marjan
  Ghazvininejad, Mike Lewis, and Luke Zettlemoyer. 2020.
\newblock \href {http://arxiv.org/abs/2001.08210} {Multilingual denoising
  pre-training for neural machine translation}.

\bibitem[{Liu et~al.(2019{\natexlab{b}})Liu, Ott, Goyal, Du, Joshi, Chen, Levy,
  Lewis, Zettlemoyer, and Stoyanov}]{liu2019roberta}
Yinhan Liu, Myle Ott, Naman Goyal, Jingfei Du, Mandar Joshi, Danqi Chen, Omer
  Levy, Mike Lewis, Luke Zettlemoyer, and Veselin Stoyanov. 2019{\natexlab{b}}.
\newblock \href {http://arxiv.org/abs/1907.11692} {Roberta: A robustly
  optimized bert pretraining approach}.

\bibitem[{Lugosch et~al.(2019)Lugosch, Ravanelli, Ignoto, Tomar, and
  Bengio}]{lugosch2019speech}
Loren Lugosch, Mirco Ravanelli, Patrick Ignoto, Vikrant~Singh Tomar, and Yoshua
  Bengio. 2019.
\newblock \href {https://doi.org/10.21437/Interspeech.2019-2396} {{Speech Model
  Pre-Training for End-to-End Spoken Language Understanding}}.
\newblock In \emph{Proc. Interspeech 2019}, pages 814--818.

\bibitem[{Magueresse et~al.(2020)Magueresse, Carles, and
  Heetderks}]{magueresse2020lowresource}
Alexandre Magueresse, Vincent Carles, and Evan Heetderks. 2020.
\newblock \href {http://arxiv.org/abs/2006.07264} {Low-resource languages: A
  review of past work and future challenges}.

\bibitem[{Marivate et~al.(2020)Marivate, Sefara, Chabalala, Makhaya,
  Mokgonyane, Mokoena, and Modupe}]{marivate-etal-2020-investigating}
Vukosi Marivate, Tshephisho Sefara, Vongani Chabalala, Keamogetswe Makhaya,
  Tumisho Mokgonyane, Rethabile Mokoena, and Abiodun Modupe. 2020.
\newblock \href {https://aclanthology.org/2020.rail-1.3} {Investigating an
  approach for low resource language dataset creation, curation and
  classification: Setswana and sepedi}.
\newblock In \emph{Proceedings of the first workshop on Resources for African
  Indigenous Languages}, pages 15--20, Marseille, France. European Language
  Resources Association (ELRA).

\bibitem[{Mayhew et~al.(2017)Mayhew, Tsai, and Roth}]{mayhew-etal-2017-cheap}
Stephen Mayhew, Chen-Tse Tsai, and Dan Roth. 2017.
\newblock \href {https://doi.org/10.18653/v1/D17-1269} {Cheap translation for
  cross-lingual named entity recognition}.
\newblock In \emph{Proceedings of the 2017 Conference on Empirical Methods in
  Natural Language Processing}, pages 2536--2545, Copenhagen, Denmark.
  Association for Computational Linguistics.

\bibitem[{Nicosia et~al.(2021)Nicosia, Qu, and
  Altun}]{nicosia-etal-2021-translate-fill}
Massimo Nicosia, Zhongdi Qu, and Yasemin Altun. 2021.
\newblock \href {https://doi.org/10.18653/v1/2021.findings-emnlp.279}
  {{T}ranslate {\&} {F}ill: {I}mproving zero-shot multilingual semantic parsing
  with synthetic data}.
\newblock In \emph{Findings of the Association for Computational Linguistics:
  EMNLP 2021}, pages 3272--3284, Punta Cana, Dominican Republic. Association
  for Computational Linguistics.

\bibitem[{Pan et~al.(2017)Pan, Zhang, May, Nothman, Knight, and
  Ji}]{Pan2017CrosslingualNT}
Xiaoman Pan, Boliang Zhang, Jonathan May, Joel Nothman, Kevin Knight, and Heng
  Ji. 2017.
\newblock Cross-lingual name tagging and linking for 282 languages.
\newblock In \emph{ACL}.

\bibitem[{Pires et~al.(2019)Pires, Schlinger, and
  Garrette}]{pires-etal-2019-multilingual}
Telmo Pires, Eva Schlinger, and Dan Garrette. 2019.
\newblock \href {https://doi.org/10.18653/v1/P19-1493} {How multilingual is
  multilingual {BERT}?}
\newblock In \emph{Proceedings of the 57th Annual Meeting of the Association
  for Computational Linguistics}, pages 4996--5001, Florence, Italy.
  Association for Computational Linguistics.

\bibitem[{Price(1990)}]{price-1990-evaluation}
P.~J. Price. 1990.
\newblock \href {https://aclanthology.org/H90-1020} {Evaluation of spoken
  language systems: the {ATIS} domain}.
\newblock In \emph{Speech and Natural Language: Proceedings of a Workshop Held
  at Hidden Valley, {P}ennsylvania, June 24-27,1990}.

\bibitem[{Rajpurkar et~al.(2016)Rajpurkar, Zhang, Lopyrev, and
  Liang}]{rajpurkar2016squad}
Pranav Rajpurkar, Jian Zhang, Konstantin Lopyrev, and Percy Liang. 2016.
\newblock \href {http://arxiv.org/abs/1606.05250} {Squad: 100,000+ questions
  for machine comprehension of text}.

\bibitem[{Ronen et~al.(2014)Ronen, Gonçalves, Hu, Vespignani, Pinker, and
  Hidalgo}]{links-that-speak}
Shahar Ronen, Bruno Gonçalves, Kevin~Z. Hu, Alessandro Vespignani, Steven
  Pinker, and César~A. Hidalgo. 2014.
\newblock \href {https://doi.org/10.1073/pnas.1410931111} {Links that speak:
  The global language network and its association with global fame}.
\newblock \emph{Proceedings of the National Academy of Sciences},
  111(52):E5616--E5622.

\bibitem[{Rongali et~al.(2020)Rongali, Soldaini, Monti, and
  Hamza}]{dontparsegenerate}
Subendhu Rongali, Luca Soldaini, Emilio Monti, and Wael Hamza. 2020.
\newblock \href {https://doi.org/10.1145/3366423.3380064} {Don’t parse,
  generate! a sequence to sequence architecture for task-oriented semantic
  parsing}.
\newblock \emph{Proceedings of The Web Conference 2020}.

\bibitem[{Saade et~al.(2019)Saade, Coucke, Caulier, Dureau, Ball, Bluche,
  Leroy, Doumouro, Gisselbrecht, Caltagirone, Lavril, and
  Primet}]{saade2019spoken}
Alaa Saade, Alice Coucke, Alexandre Caulier, Joseph Dureau, Adrien Ball,
  Th{\'e}odore Bluche, David Leroy, Cl{\'e}ment Doumouro, Thibault
  Gisselbrecht, Francesco Caltagirone, Thibaut Lavril, and Ma{\"e}l Primet.
  2019.
\newblock \href {http://arxiv.org/abs/1810.12735} {Spoken language
  understanding on the edge}.

\bibitem[{Schuster et~al.(2019)Schuster, Gupta, Shah, and
  Lewis}]{schuster-etal-2019-cross-lingual}
Sebastian Schuster, Sonal Gupta, Rushin Shah, and Mike Lewis. 2019.
\newblock \href {https://doi.org/10.18653/v1/N19-1380} {Cross-lingual transfer
  learning for multilingual task oriented dialog}.
\newblock In \emph{Proceedings of the 2019 Conference of the North {A}merican
  Chapter of the Association for Computational Linguistics: Human Language
  Technologies, Volume 1 (Long and Short Papers)}, pages 3795--3805,
  Minneapolis, Minnesota. Association for Computational Linguistics.

\bibitem[{Simons(2022)}]{ethnologue}
Gary Simons, editor. 2022.
\newblock \href {https://www.ethnologue.com/} {\emph{Ethnologue: Languages of
  the World}}, twenty-fifth edition.
\newblock SIL International, Dallas, TX, USA.

\bibitem[{Simpson et~al.(2008)Simpson, Cieri, Maeda, Baker, and
  Onyshkevych}]{simpson2008human}
Heather Simpson, Christopher Cieri, Kazuaki Maeda, Kathryn Baker, and Boyan
  Onyshkevych. 2008.
\newblock Human language technology resources for less commonly taught
  languages: Lessons learned toward creation of basic language resources.
\newblock \emph{Collaboration: interoperability between people in the creation
  of language resources for less-resourced languages}, 7.

\bibitem[{Singla et~al.(2018)Singla, Can, and
  Narayanan}]{singla-etal-2018-multi}
Karan Singla, Dogan Can, and Shrikanth Narayanan. 2018.
\newblock \href {https://doi.org/10.18653/v1/P18-2035} {A multi-task approach
  to learning multilingual representations}.
\newblock In \emph{Proceedings of the 56th Annual Meeting of the Association
  for Computational Linguistics (Volume 2: Short Papers)}, pages 214--220,
  Melbourne, Australia. Association for Computational Linguistics.

\bibitem[{Strassel and Tracey(2016)}]{strassel-tracey-2016-lorelei}
Stephanie Strassel and Jennifer Tracey. 2016.
\newblock \href {https://aclanthology.org/L16-1521} {{LORELEI} language packs:
  Data, tools, and resources for technology development in low resource
  languages}.
\newblock In \emph{Proceedings of the Tenth International Conference on
  Language Resources and Evaluation ({LREC}'16)}, pages 3273--3280,
  Portoro{\v{z}}, Slovenia. European Language Resources Association (ELRA).

\bibitem[{Thattai et~al.(2020)Thattai, Tur, and
  Natarajan}]{thattai_tur_natarajan_2020}
Govind Thattai, Gokhan Tur, and Prem Natarajan. 2020.
\newblock \href
  {https://www.amazon.science/blog/new-alexa-features-interactive-teaching-by-customers}
  {New alexa features: Interactive teaching by customers}.

\bibitem[{Tiedemann(2012)}]{TIEDEMANN12.463}
Jörg Tiedemann. 2012.
\newblock Parallel data, tools and interfaces in opus.
\newblock In \emph{Proceedings of the Eight International Conference on
  Language Resources and Evaluation (LREC'12)}, Istanbul, Turkey. European
  Language Resources Association (ELRA).

\bibitem[{Tur et~al.(2010)Tur, Hakkani-T{\"u}r, and Heck}]{tur2010left}
Gokhan Tur, Dilek Hakkani-T{\"u}r, and Larry Heck. 2010.
\newblock What is left to be understood in atis?
\newblock In \emph{2010 IEEE Spoken Language Technology Workshop}, pages
  19--24. IEEE.

\bibitem[{Tur and Mori(2011)}]{Tur2011SpokenLU}
Gokhan Tur and Renato~De Mori. 2011.
\newblock Spoken language understanding: Systems for extracting semantic
  information from speech.

\bibitem[{Upadhyay et~al.(2018)Upadhyay, Faruqui, T{\"u}r, Dilek, and
  Heck}]{8461905}
Shyam Upadhyay, Manaal Faruqui, Gokhan T{\"u}r, Hakkani-T{\"u}r Dilek, and
  Larry Heck. 2018.
\newblock \href {https://doi.org/10.1109/ICASSP.2018.8461905} {(almost)
  zero-shot cross-lingual spoken language understanding}.
\newblock In \emph{2018 IEEE International Conference on Acoustics, Speech and
  Signal Processing (ICASSP)}, pages 6034--6038.

\bibitem[{Wang et~al.(2021)Wang, Gaspers, Do, and
  Jiang}]{wang-etal-2021-exploring-cross}
Chao Wang, Judith Gaspers, Thi Ngoc~Quynh Do, and Hui Jiang. 2021.
\newblock \href {https://doi.org/10.18653/v1/2021.findings-acl.177} {Exploring
  cross-lingual transfer learning with unsupervised machine translation}.
\newblock In \emph{Findings of the Association for Computational Linguistics:
  ACL-IJCNLP 2021}, pages 2011--2020, Online. Association for Computational
  Linguistics.

\bibitem[{Wang et~al.(2005)Wang, Deng, and Acero}]{Wang2005SpokenLU}
Ye-Yi Wang, Li~Deng, and Alex Acero. 2005.
\newblock Spoken language understanding.
\newblock \emph{IEEE Signal Processing Magazine}, 22:16--31.

\bibitem[{Wolf et~al.(2020)Wolf, Debut, Sanh, Chaumond, Delangue, Moi, Cistac,
  Rault, Louf, Funtowicz, Davison, Shleifer, von Platen, Ma, Jernite, Plu, Xu,
  Scao, Gugger, Drame, Lhoest, and Rush}]{wolf-etal-2020-transformers}
Thomas Wolf, Lysandre Debut, Victor Sanh, Julien Chaumond, Clement Delangue,
  Anthony Moi, Pierric Cistac, Tim Rault, Rémi Louf, Morgan Funtowicz, Joe
  Davison, Sam Shleifer, Patrick von Platen, Clara Ma, Yacine Jernite, Julien
  Plu, Canwen Xu, Teven~Le Scao, Sylvain Gugger, Mariama Drame, Quentin Lhoest,
  and Alexander~M. Rush. 2020.
\newblock \href {https://www.aclweb.org/anthology/2020.emnlp-demos.6}
  {Transformers: State-of-the-art natural language processing}.
\newblock In \emph{Proceedings of the 2020 Conference on Empirical Methods in
  Natural Language Processing: System Demonstrations}, pages 38--45, Online.
  Association for Computational Linguistics.

\bibitem[{Xie et~al.(2018)Xie, Yang, Neubig, Smith, and
  Carbonell}]{xie-etal-2018-neural}
Jiateng Xie, Zhilin Yang, Graham Neubig, Noah~A. Smith, and Jaime Carbonell.
  2018.
\newblock \href {https://doi.org/10.18653/v1/D18-1034} {Neural cross-lingual
  named entity recognition with minimal resources}.
\newblock In \emph{Proceedings of the 2018 Conference on Empirical Methods in
  Natural Language Processing}, pages 369--379, Brussels, Belgium. Association
  for Computational Linguistics.

\bibitem[{Xu et~al.(2020)Xu, Haider, and Mansour}]{xu-etal-2020-end}
Weijia Xu, Batool Haider, and Saab Mansour. 2020.
\newblock \href {https://doi.org/10.18653/v1/2020.emnlp-main.410} {End-to-end
  slot alignment and recognition for cross-lingual {NLU}}.
\newblock In \emph{Proceedings of the 2020 Conference on Empirical Methods in
  Natural Language Processing (EMNLP)}, pages 5052--5063, Online. Association
  for Computational Linguistics.

\bibitem[{Xue et~al.(2021)Xue, Constant, Roberts, Kale, Al-Rfou, Siddhant,
  Barua, and Raffel}]{xue-etal-2021-mt5}
Linting Xue, Noah Constant, Adam Roberts, Mihir Kale, Rami Al-Rfou, Aditya
  Siddhant, Aditya Barua, and Colin Raffel. 2021.
\newblock \href {https://doi.org/10.18653/v1/2021.naacl-main.41} {m{T}5: A
  massively multilingual pre-trained text-to-text transformer}.
\newblock In \emph{Proceedings of the 2021 Conference of the North American
  Chapter of the Association for Computational Linguistics: Human Language
  Technologies}, pages 483--498, Online. Association for Computational
  Linguistics.

\bibitem[{Yarowsky et~al.(2001)Yarowsky, Ngai, and
  Wicentowski}]{yarowsky-etal-2001-inducing}
David Yarowsky, Grace Ngai, and Richard Wicentowski. 2001.
\newblock \href {https://aclanthology.org/H01-1035} {Inducing multilingual text
  analysis tools via robust projection across aligned corpora}.
\newblock In \emph{Proceedings of the First International Conference on Human
  Language Technology Research}.

\bibitem[{Young(2002)}]{Young2002TalkingTM}
Steve~J. Young. 2002.
\newblock Talking to machines (statistically speaking).
\newblock In \emph{INTERSPEECH}.

\bibitem[{Zhu et~al.(2019)Zhu, Zhao, Zhao, Zong, and
  Yu}]{10.1145/3340555.3356098}
Su~Zhu, Zijian Zhao, Tiejun Zhao, Chengqing Zong, and Kai Yu. 2019.
\newblock \href {https://doi.org/10.1145/3340555.3356098} {Catslu: The 1st
  chinese audio-textual spoken language understanding challenge}.
\newblock In \emph{2019 International Conference on Multimodal Interaction},
  ICMI '19, pages 521--525, New York, NY, USA. Association for Computing
  Machinery.

\end{thebibliography}
\balance
\bibliographystyle{acl_natbib}

\clearpage

\appendix

\section{Additional Linguistic Characteristics}

Additional linguistic characteristics of our languages are given in Table \ref{tab:addl_ling_char}.

\begin{table*}[]
\resizebox{\linewidth}{!}{%
\begin{tabular}{lllllllllllll}
\toprule
Name & Code & WALS & \makecell{ISO\\639-3} & Family & Subdivision & Script & Order & Politeness & Imperative Morphology & \makecell{Imperative\\Hortative} & Optative & Prohibitive \\
\midrule
Afrikaans & af-ZA & afr & afr & Indo-European & Germanic & Latin & - & - & - & - & - & - \\
Albanian & sq-AL & alb & aln & Indo-European & Albanian & Latin & SVO & None & singular only & minimal & present & special negative \\
Amharic & am-ET & amh & amh & Afro-Asiatic & Semtic & Ge'ez & SOV & - & singular and plural & neither & - & special imperative and negative \\
Arabic & ar-SA & ams & arb & Afro-Asiatic & Semtic & Arabic & VSO & - & - & - & - & - \\
Armenian & hy-AM & arm & hye & Indo-European & Armenian & Armenian & None & binary & singular and plural & neither & absent & special negative \\
Azerbaijani & az-AZ & aze & azj & Turkic & Oghuz & Latin & SOV & - & - & - & present & - \\
Bengali & bn-BD & ben & ben & Indo-European & Indo-Aryan & Bengali & SOV & - & - & - & - & - \\
Burmese & my-MM & brm & mya & Sino-Tibetan & Lolo-Burmese & Burmese & SOV & avoidance & None & neither & absent & special negative \\
Danish & da-DK & dsh & dan & Indo-European & Germanic & Latin & SVO & binary & number neutral & neither & absent & normal imperative and negative \\
Dutch & nl-NL & dut & nld & Indo-European & Germanic & Latin & None & binary & number neutral & neither & - & normal imperative and negative \\
English & en-US & eng & eng & Indo-European & Germanic & Latin & SVO & None & None & neither & absent & normal imperative and negative \\
Finnish & fi-FI & fin & fin & Uralic & Finno-Ugric & Latin & SVO & binary & singular and plural & minimal & absent & special negative \\
French & fr-FR & fre & fra & Indo-European & Romance & Latin & SVO & binary & singular only & neither & absent & normal imperative and negative \\
Georgian & ka-GE & geo & kat & Kartvelian & Karto-Zan & Georgian & SOV & binary & None & neither & present & - \\
German & de-DE & ger & deu & Indo-European & Germanic & Latin & None & binary & singular only & neither & absent & normal imperative and negative \\
Greek & el-GR & grk & ell & Indo-European & Hellenic & Greek & None & binary & singular and plural & minimal & absent & special imperative and negative \\
Hebrew & he-IL & heb & heb & Afro-Asiatic & Semtic & Hebrew & SVO & None & singular and plural & minimal & absent & special imperative and negative \\
Hindi & hi-IN & hin & hin & Indo-European & Indo-Aryan & Devanagari & SOV & multiple & singular and plural & neither & absent & special negative \\
Hungarian & hu-HU & hun & hun & Uralic & Finno-Ugric & Latin & None & multiple & None & minimal & absent & special negative \\
Icelandic & is-IS & ice & isl & Indo-European & Germanic & Latin & SVO & - & singular only & neither & absent & normal imperative and negative \\
Indonesian & id-ID & ind & ind & Austronesian & Malayic & Latin & SVO & avoidance & None & neither & absent & special negative \\
Italian & it-IT & ita & ita & Indo-European & Romance & Latin & SVO & binary & singular only & neither & - & special imperative \\
Japanese & ja-JP & jpn & jpn & Japonic & Japanese & Japanese & SOV & avoidance & number neutral & neither & absent & special negative \\
Javanese & jv-ID & jav & jav & Austronesian & Javanese & Latin & - & - & - & neither & - & - \\
Kannada & kn-IN & knd & kan & Dravidian & Southern & Kannada & SOV & multiple & singular and plural & minimal & absent & special imperative and negative \\
Khmer & km-KH & khm & khm & Austoasiatic & Khmeric & Khmer & SVO & avoidance & None & - & absent & special negative \\
Korean & ko-KR & kor & kor & Koreanic & Korean & Hangul & SOV & avoidance & number neutral & neither & absent & special negative \\
Latvian & lv-LV & lat & lav & Indo-European & Baltic & Latin & SVO & binary & plural only & neither & absent & normal imperative and negative \\
Malay & ms-MY & mly & zsm & Austronesian & Malayic & Latin & - & - & - & - & - & - \\
Malayalam & ml-IN & mym & mal & Dravidian & Southern & Malayalam & SOV & multiple & singular and plural & neither & absent & special negative \\
Mandarin (simp) & zh-CN & mnd & cmn & Sino-Tibetan & Sinitic & Simp Chinese & SVO & binary & None & neither & absent & special negative \\
Mandarin (trad) & zh-TW & mnd & cmn & Sino-Tibetan & Sinitic & Trad Chinese & SVO & binary & None & neither & absent & special negative \\
Mongolian & mn-MN & - & mon & Mongolic & Mongolian & Cyrillic & - & - & - & - & - & - \\
Norwegian & nb-NO & nor & nob & Indo-European & Germanic & Latin & SVO & binary & number neutral & neither & absent & normal imperative and negative \\
Persian & fa-IR & prs & pes & Indo-European & Indo-Iranian & Arabic & SOV & binary & singular only & maximal & absent & normal imperative and negative \\
Polish & pl-PL & pol & pol & Indo-European & Slavic & Latin & SVO & binary & singular and plural & neither & - & normal imperative and negative \\
Portuguese & pt-PT & por & por & Indo-European & Romance & Latin & SVO & binary & singular only & neither & - & special imperative \\
Romanian & ro-RO & rom & ron & Indo-European & Romance & Latin & SVO & multiple & singular only & minimal & - & special imperative \\
Russian & ru-RU & rus & rus & Indo-European & Slavic & Cyrillic & SVO & binary & singular and plural & neither & absent & normal imperative and negative \\
Slovenian & sl-SI & slo & slv & Indo-European & Slavic & Latin & SVO & - & singular and plural & neither & absent & normal imperative and negative \\
Spanish & es-ES & spa & spa & Indo-European & Romance & Latin & SVO & binary & singular and plural & neither & absent & special imperative \\
Swahili & sw-KE & swa & swh & Niger-Congo & Bantu & Latin & SVO & None & singular and plural & minimal & absent & special imperative and negative \\
Swedish & sv-SE & swe & swe & Indo-European & Germanic & Latin & SVO & binary & number neutral & neither & absent & normal imperative and negative \\
Tagalog & tl-PH & tag & tgl & Austronesian & Philippine & Latin & VSO & multiple & singular and plural & neither & present & special negative \\
Tamil & ta-IN & tml & tam & Dravidian & Southern & Tamil & SOV & multiple & singular and plural & - & - & special imperative and negative \\
Telugu & te-IN & tel & tel & Dravidian & South-Central & Telugu & SOV & - & singular and plural & - & absent & special negative \\
Thai & th-TH & tha & tha & Kra-Dai & Tai & Thai & SVO & avoidance & None & neither & absent & special negative \\
Turkish & tr-TR & tur & tur & Turkic & Oghuz & Latin & SOV & binary & singular and plural & minimal & absent & normal imperative and negative \\
Urdu & ur-PK & urd & urd & Indo-European & Indo-Aryan & Arabic & SOV & multiple & - & - & absent & - \\
Vietnamese & vi-VN & vie & vie & Austoasiatic & Vietic & Latin & SVO & avoidance & None & neither & absent & special negative \\
Welsh & cy-GB & wel & cym & Indo-European & Celtic & Latin & VSO & binary & singular and plural & neither & - & special negative \\
\bottomrule
\end{tabular}
}
\caption{Additional linguistic characteristics of the \M{} languages.}
\label{tab:addl_ling_char}
\end{table*}

\section{The Collection System}
\label{sec:appendix}

\begin{figure*}
  \centering \includegraphics[width=\textwidth]{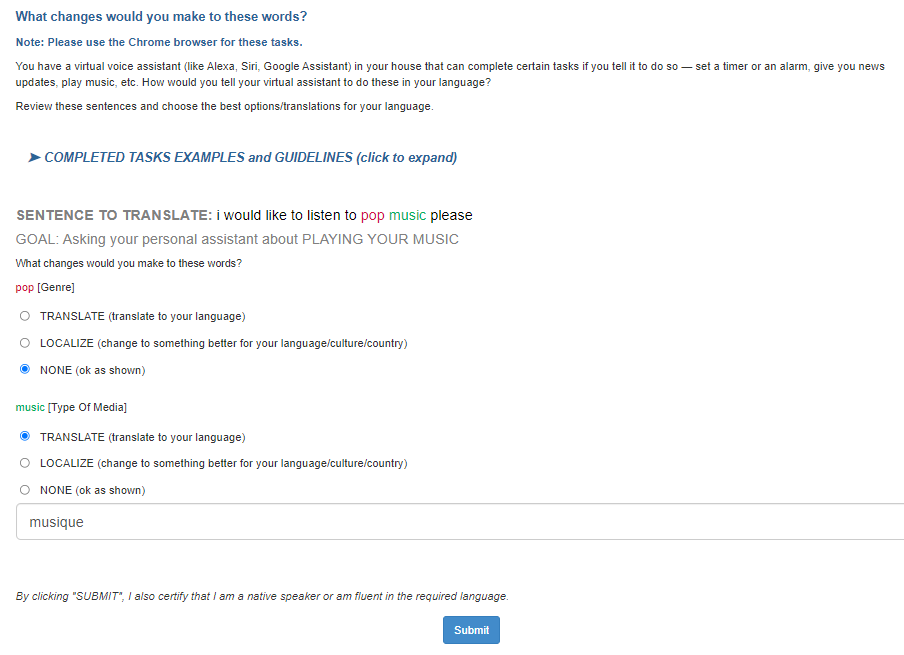}
  \caption{Slot localization task as presented to Amazon MTurk worker.} \label{fig:slot}
\end{figure*}

\begin{figure*}
  \centering \includegraphics[width=\textwidth]{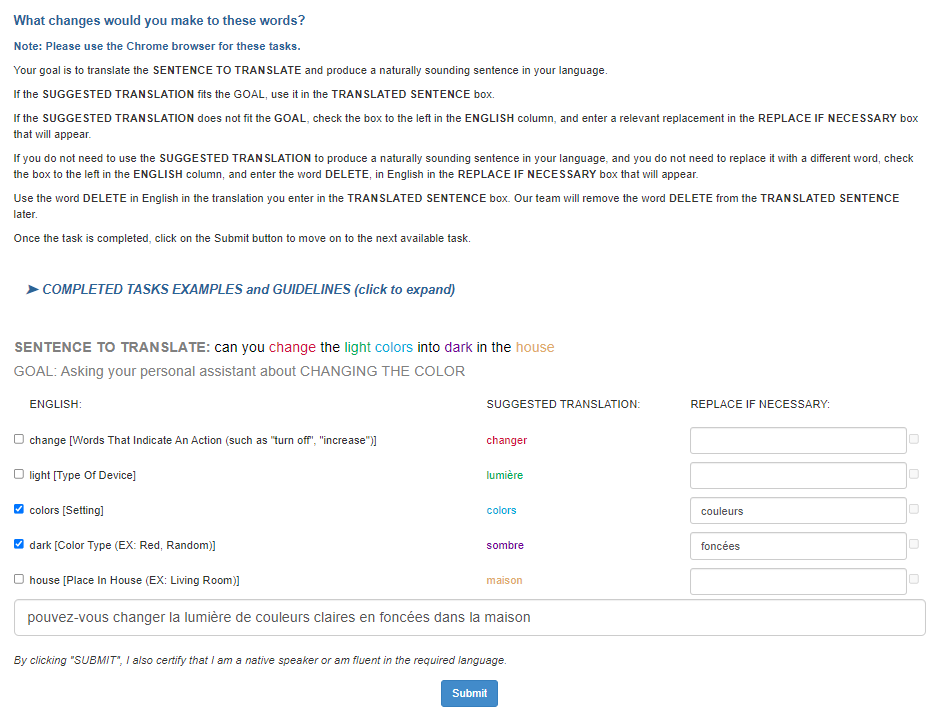}
  \caption{Phrase localization task as presented to Amazon MTurk worker.} \label{fig:phrase}
\end{figure*}

\begin{figure*}
  \centering \includegraphics[width=\textwidth]{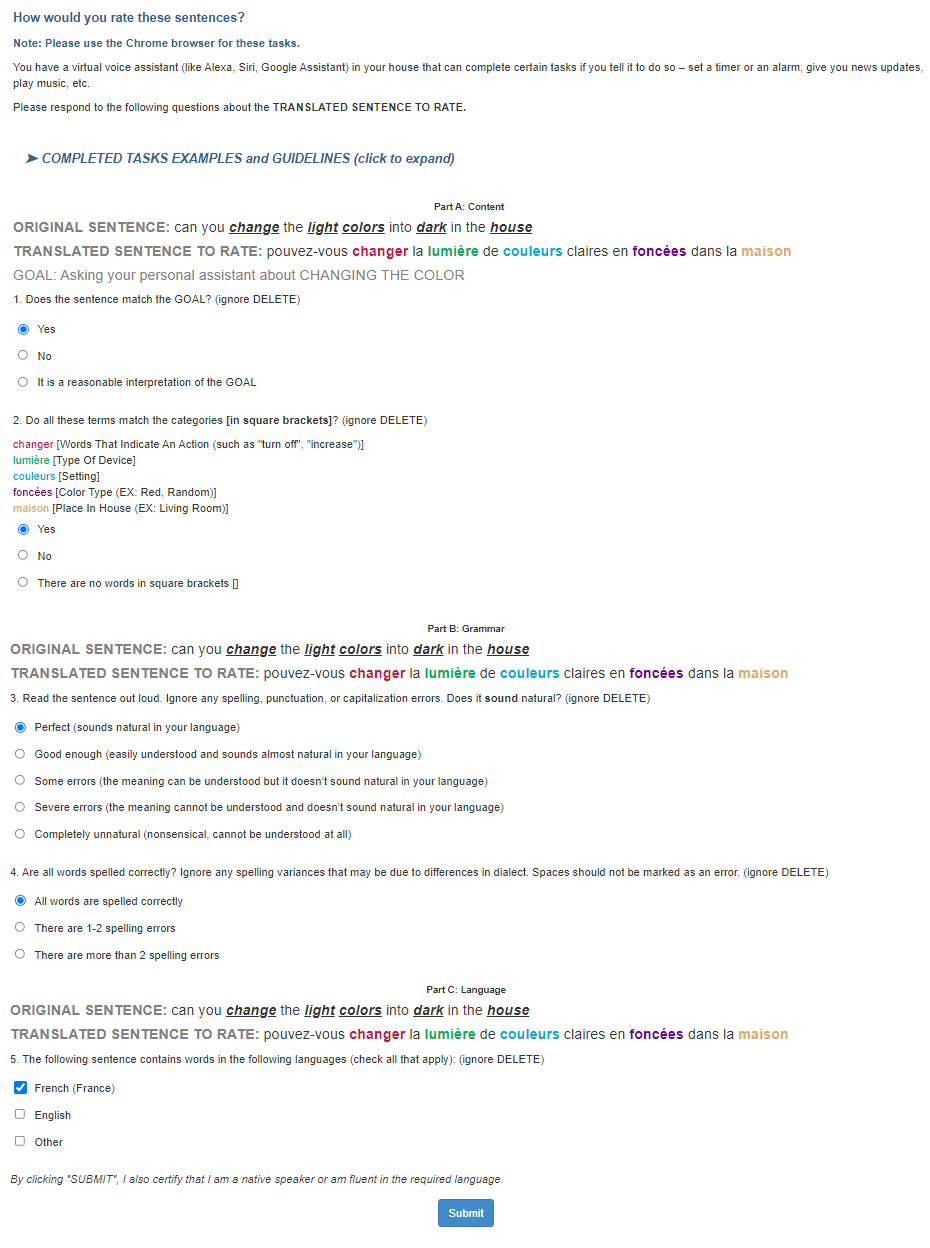}
  \caption{Judgment task as presented to Amazon MTurk worker.} \label{fig:judgment}
\end{figure*}

Screenshots from our collection workflow are given in Figures \ref{fig:slot}, \ref{fig:phrase}, and \ref{fig:judgment}.

\section{Hyperparameters}

The hyperparameter search spaces and the chosen hyperparameters are given in Tables \ref{tab:model_hyperparams} and \ref{tab:model_hyperparams_zero}.

\begin{table*}[]
\resizebox{\linewidth}{!}{%
\begin{tabular}{lccc}
\toprule
  & XLM-R Base & mT5 Text-to-Text & mT5 Encoder-Only \\ 
\midrule
Adam $\beta_1$  & [0.8, 0.9, 0.99] & [0.8, 0.9, 0.99] & [0.8, 0.9, 0.99] \\ 
  & choice & choice & choice \\ 
  & 0.9 & 0.8 & 0.8 \\ 
Adam $\beta_2$  & [0.95, 0.99, 0.999, 0.9999] & [0.95, 0.99, 0.999, 0.9999] & [0.95, 0.99, 0.999, 0.9999] \\ 
  & choice & choice & choice \\ 
  & 0.9999 & 0.9999 & 0.999 \\ 
Adam $\epsilon$  & [1e-06, 1e-07, 1e-08, 1e-09] & [1e-06, 1e-07, 1e-08, 1e-09] & [1e-06, 1e-07, 1e-08, 1e-09] \\ 
  & choice & choice & choice \\ 
  & 1e-08 & 1e-09 & 1e-09 \\ 
Batch Size  & [32, 64, 128, 256, 512, 1024] & [8, 16, 32, 64] &  \\ 
  & choice & choice &  \\ 
  & 1024 & 64 &  \\ 
Dropout, Attention  & [0.0, 0.5, 0.05] &  & [0.0, 0.5, 0.05] \\ 
  & quniform &  & quniform \\ 
  & 0.0 &  & 0.45 \\ 
Dropout, Feedforward  & [0.0, 0.5, 0.05] & [0.0, 0.5, 0.05] & [0.0, 0.5, 0.05] \\ 
  & quniform & quniform & quniform \\ 
  & 0.45 & 0.05 & 0.25 \\ 
Encoder Layer Used  & [7, 8, 9, 10, 11] &  & [7, 8, 9, 10, 11] \\ 
  & choice &  & choice \\ 
  & 11 &  & 9 \\ 
Generation Num Beams  &  & [1, 2, 3] &  \\ 
  &  & choice &  \\ 
  &  & 2 &  \\ 
Gradient Accumulation Steps  &  &  & [4, 8, 16, 32, 64]\\ 
  &  &  & choice  \\ 
  &  &  & 64 \\ 
Hid Dim Class Head  & [256, 512, 728, 1024, 2048] &  & [256, 512, 728, 1024, 2048] \\ 
  & choice &  & choice \\ 
  & 2048 &  & 1024 \\ 
Intent Class Pooling  & [first, max, mean] &  & [first, max, mean] \\ 
  & choice &  & choice \\ 
  & max &  & first \\ 
LR Scheduler  & [linear, constant\_with\_warmup] & [linear, constant\_with\_warmup] & [linear, constant\_with\_warmup] \\ 
  & choice & choice & choice \\ 
  & constant\_with\_warmup & linear & constant\_with\_warmup \\ 
Learning Rate  & [1e-07, 0.0001, 1e-07] & [1e-07, 0.001, 1e-07] & [1e-07, 0.001, 1e-07] \\ 
  & qloguniform & qloguniform & qloguniform \\ 
  & 2.8e-05 & 8e-05 & 0.0003525 \\ 
Num Layers Class Head  & [0, 1, 2, 3] &  & [0, 1, 2, 3] \\ 
  & choice &  & choice \\ 
  & 1 &  & 1 \\ 
Slot Loss Coefficient  & [0.5, 1.0, 2.0, 4.0, 8.0, 16.0] &  & [0.5, 1.0, 2.0, 4.0, 8.0, 16.0] \\ 
  & choice &  & choice \\ 
  & 4.0 &  & 4.0 \\ 
Tot Epochs, LR Sched  & [3, 30, 1] & [3, 30, 1] & [3, 30, 1] \\ 
  & quniform & quniform & quniform \\ 
  & 26 & 22 & 15 \\ 
Warmup Steps  & [0, 1000, 100] & [0, 1000, 100] & [0, 1000, 100] \\ 
  & quniform & quniform & quniform \\ 
  & 800 & 200 & 600 \\ 
Weight Decay  & [0.0, 0.5, 0.01] & [0.0, 0.5, 0.01] & [0.0, 0.5, 0.01] \\ 
  & quniform & quniform & quniform \\ 
  & 0.21 & 0.16 & 0.07 \\ 
\bottomrule
\end{tabular}
}
\caption{The full-dataset hyperparameter search space, the sampling technique, and the chosen hyperparameter for our 3 models. The search space for the ``quniform'' and ``qloguniform'' sampling techniques is given as [min, max, increment].}
\label{tab:model_hyperparams}
\end{table*}

\begin{table*}[]
\resizebox{\linewidth}{!}{%
\begin{tabular}{lccc}
\toprule
  & XLM-R Base & mT5 Text-to-Text & mT5 Encoder-Only \\ 
\midrule
Adam $\beta_1$  & [0.8, 0.9, 0.99] & [0.8, 0.9, 0.99] & [0.8, 0.9, 0.99] \\ 
  & choice & choice & choice \\ 
  & 0.99 & 0.8 & 0.8 \\ 
Adam $\beta_2$  & [0.95, 0.99, 0.999, 0.9999] & [0.95, 0.99, 0.999, 0.9999] & [0.95, 0.99, 0.999, 0.9999] \\ 
  & choice & choice & choice \\ 
  & 0.9999 & 0.999 & 0.9999 \\ 
Adam $\epsilon$  & [1e-06, 1e-07, 1e-08, 1e-09] & [1e-06, 1e-07, 1e-08, 1e-09] & [1e-06, 1e-07, 1e-08, 1e-09] \\ 
  & choice & choice & choice \\ 
  & 1e-09 & 1e-09 & 1e-08 \\ 
Batch Size  &  &  &  \\ 
  &  &  &  \\ 
  &  &  &  \\ 
Dropout, Attention  & [0.0, 0.5, 0.05] &  & [0.0, 0.5, 0.05] \\ 
  & quniform &  & quniform \\ 
  & 0.35 &  & 0.4 \\ 
Dropout, Feedforward  & [0.0, 0.5, 0.05] & [0.0, 0.5, 0.05] & [0.0, 0.5, 0.05] \\ 
  & quniform & quniform & quniform \\ 
  & 0.25 & 0.2 & 0.2 \\ 
Encoder Layer Used  & [7, 8, 9, 10, 11] &  & [7, 8, 9, 10, 11] \\ 
  & choice &  & choice \\ 
  & 10 &  & 8 \\ 
Freeze Layers  & [xlmr.embeddings.word\_embeddings.weight, & [shared.weight, & [mt5.shared.weight, \\ 
                         &            null]         &   shared.weight + lm\_head.weight, null]     & null]  \\
  & choice & choice & choice \\ 
  & xlmr.embeddings.word\_embeddings.weight & null & mt5.shared.weight \\
Generation Num Beams  &  & [1, 2, 3] &  \\ 
  &  & choice &  \\ 
  &  & 3 &  \\ 
Gradient Accumulation Steps  & [1, 2, 4, 8, 16, 32] & [4, 8, 16, 32, 64] & [4, 8, 16, 32, 64]\\ 
  & choice & choice & choice  \\ 
  & 8 & 64 & 32 \\ 
Hid Dim Class Head  & [728, 1024, 2048, 3072, 4096, 8192, 16384] &  & [256, 512, 728, 1024, 2048] \\ 
  & choice &  & choice \\ 
  & 8192 &  & 2048 \\ 
Intent Class Pooling  & [first, max, mean] &  & [first, max, mean] \\ 
  & choice &  & choice \\ 
  & max &  & mean \\ 
LR Scheduler  & [linear, constant\_with\_warmup] & [linear, constant\_with\_warmup] & [linear, constant\_with\_warmup] \\ 
  & choice & choice & choice \\ 
  & constant\_with\_warmup & linear & linear \\ 
Learning Rate  & [1e-07, 0.0001, 1e-07] & [1e-07, 0.001, 1e-07] & [1e-07, 0.001, 1e-07] \\ 
  & qloguniform & qloguniform & qloguniform \\ 
  & 4.7e-06 & 3.4e-05 & 6.19e-05 \\ 
Num Layers Class Head  & [0, 1, 2, 3] &  & [0, 1, 2, 3] \\ 
  & choice &  & choice \\ 
  & 2 &  & 3 \\ 
Slot Loss Coefficient  & [0.5, 1.0, 2.0, 4.0, 8.0, 16.0] &  & [0.5, 1.0, 2.0, 4.0, 8.0, 16.0] \\ 
  & choice &  & choice \\ 
  & 2.0 &  & 4.0 \\ 
Tot Epochs, LR Sched  & [50, 1500, 50] & [50, 1500, 50] & [30, 1500, 10] \\ 
  & quniform & quniform & quniform \\ 
  & 850 & 950 & 300 \\ 
Warmup Steps  & [0, 1000, 100] & [0, 1000, 100] & [0, 1000, 100] \\ 
  & quniform & quniform & quniform \\ 
  & 500 & 300 & 700 \\ 
Weight Decay  & [0.0, 0.5, 0.01] & [0.0, 0.5, 0.01] & [0.0, 0.5, 0.01] \\ 
  & quniform & quniform & quniform \\ 
  & 0.11 & 0.0  & 0.35 \\ 
\bottomrule
\end{tabular}
}
\caption{The zero-shot hyperparameter search space, the sampling technique, and the chosen hyperparameter for our 3 models. The search space for the ``quniform'' and ``qloguniform'' sampling techniques is given as [min, max, increment].}
\label{tab:model_hyperparams_zero}
\end{table*}

\section{Results for All Languages}
\label{sect:results_all_langs}
Results for all languages are given for exact match accuracy in Table \ref{tab:all_ex_match}, intent accuracy in Table \ref{tab:all_intent_acc}, and micro-averaged slot-filling F1 in Table \ref{tab:all_slot_f1}.

\begin{table*}[]
\centering
\resizebox{0.85\linewidth}{!}{%
\begin{tabular}{lcccccc}
\multicolumn{7}{c}{Exact Match Accuracy (\%)} \\
\toprule
 & mT5 T2T Full & mT5 Enc Full & XLM-R Full & mT5 T2T Zero & mT5 Enc Zero & XLM-R Zero \\
\midrule
th-TH & 73.4 $\pm$ 1.6 & 72.3 $\pm$ 1.6 & 70.1 $\pm$ 1.6 & 33.5 $\pm$ 1.7 & 40.8 $\pm$ 1.8 & 46.3 $\pm$ 1.8 \\
en-US & 72.5 $\pm$ 1.6 & 72.0 $\pm$ 1.6 & 69.7 $\pm$ 1.7 &  & & \\
sv-SE & 71.2 $\pm$ 1.6 & 70.6 $\pm$ 1.6 & 69.7 $\pm$ 1.7 & 53.2 $\pm$ 1.8 & 44.3 $\pm$ 1.8 & 57.9 $\pm$ 1.8 \\
da-DK & 70.2 $\pm$ 1.6 & 70.3 $\pm$ 1.6 & 68.2 $\pm$ 1.7 & 47.6 $\pm$ 1.8 & 41.0 $\pm$ 1.8 & 54.4 $\pm$ 1.8 \\
my-MM & 70.1 $\pm$ 1.6 & 69.4 $\pm$ 1.7 & 65.5 $\pm$ 1.7 & 24.4 $\pm$ 1.5 & 22.2 $\pm$ 1.5 & 33.1 $\pm$ 1.7 \\
nb-NO & 70.0 $\pm$ 1.6 & 68.8 $\pm$ 1.7 & 66.8 $\pm$ 1.7 & 48.5 $\pm$ 1.8 & 41.0 $\pm$ 1.8 & 53.7 $\pm$ 1.8 \\
nl-NL & 69.4 $\pm$ 1.7 & 68.1 $\pm$ 1.7 & 66.6 $\pm$ 1.7 & 52.4 $\pm$ 1.8 & 41.0 $\pm$ 1.8 & 51.7 $\pm$ 1.8 \\
ru-RU & 69.2 $\pm$ 1.7 & 67.2 $\pm$ 1.7 & 66.2 $\pm$ 1.7 & 50.5 $\pm$ 1.8 & 42.6 $\pm$ 1.8 & 52.8 $\pm$ 1.8 \\
fi-FI & 69.1 $\pm$ 1.7 & 68.8 $\pm$ 1.7 & 66.9 $\pm$ 1.7 & 41.3 $\pm$ 1.8 & 35.8 $\pm$ 1.7 & 49.8 $\pm$ 1.8 \\
ms-MY & 69.1 $\pm$ 1.7 & 67.3 $\pm$ 1.7 & 65.6 $\pm$ 1.7 & 39.3 $\pm$ 1.8 & 33.1 $\pm$ 1.7 & 45.5 $\pm$ 1.8 \\
de-DE & 69.0 $\pm$ 1.7 & 68.9 $\pm$ 1.7 & 65.7 $\pm$ 1.7 & 52.0 $\pm$ 1.8 & 40.0 $\pm$ 1.8 & 45.4 $\pm$ 1.8 \\
ko-KR & 68.8 $\pm$ 1.7 & 68.0 $\pm$ 1.7 & 67.5 $\pm$ 1.7 & 33.7 $\pm$ 1.7 & 24.1 $\pm$ 1.5 & 44.8 $\pm$ 1.8 \\
ro-RO & 68.6 $\pm$ 1.7 & 65.1 $\pm$ 1.7 & 64.5 $\pm$ 1.7 & 45.4 $\pm$ 1.8 & 35.7 $\pm$ 1.7 & 51.6 $\pm$ 1.8 \\
id-ID & 68.6 $\pm$ 1.7 & 67.2 $\pm$ 1.7 & 64.8 $\pm$ 1.7 & 46.0 $\pm$ 1.8 & 37.4 $\pm$ 1.7 & 50.7 $\pm$ 1.8 \\
af-ZA & 68.3 $\pm$ 1.7 & 66.8 $\pm$ 1.7 & 64.9 $\pm$ 1.7 & 39.9 $\pm$ 1.8 & 34.9 $\pm$ 1.7 & 43.9 $\pm$ 1.8 \\
tr-TR & 68.1 $\pm$ 1.7 & 67.7 $\pm$ 1.7 & 65.2 $\pm$ 1.7 & 37.2 $\pm$ 1.7 & 27.4 $\pm$ 1.6 & 43.8 $\pm$ 1.8 \\
el-GR & 67.8 $\pm$ 1.7 & 66.7 $\pm$ 1.7 & 64.0 $\pm$ 1.7 & 43.5 $\pm$ 1.8 & 36.8 $\pm$ 1.7 & 41.9 $\pm$ 1.8 \\
pt-PT & 67.6 $\pm$ 1.7 & 66.0 $\pm$ 1.7 & 64.6 $\pm$ 1.7 & 47.6 $\pm$ 1.8 & 39.8 $\pm$ 1.8 & 48.6 $\pm$ 1.8 \\
hu-HU & 67.2 $\pm$ 1.7 & 67.7 $\pm$ 1.7 & 65.4 $\pm$ 1.7 & 38.7 $\pm$ 1.8 & 33.7 $\pm$ 1.7 & 44.7 $\pm$ 1.8 \\
az-AZ & 67.2 $\pm$ 1.7 & 66.2 $\pm$ 1.7 & 65.2 $\pm$ 1.7 & 28.3 $\pm$ 1.6 & 20.2 $\pm$ 1.4 & 37.2 $\pm$ 1.7 \\
is-IS & 67.1 $\pm$ 1.7 & 66.8 $\pm$ 1.7 & 64.3 $\pm$ 1.7 & 28.5 $\pm$ 1.6 & 23.4 $\pm$ 1.5 & 32.7 $\pm$ 1.7 \\
ml-IN & 67.1 $\pm$ 1.7 & 67.2 $\pm$ 1.7 & 64.9 $\pm$ 1.7 & 32.5 $\pm$ 1.7 & 27.2 $\pm$ 1.6 & 40.1 $\pm$ 1.8 \\
lv-LV & 67.0 $\pm$ 1.7 & 67.0 $\pm$ 1.7 & 66.6 $\pm$ 1.7 & 34.3 $\pm$ 1.7 & 27.4 $\pm$ 1.6 & 37.8 $\pm$ 1.7 \\
it-IT & 66.8 $\pm$ 1.7 & 64.8 $\pm$ 1.7 & 63.1 $\pm$ 1.7 & 45.1 $\pm$ 1.8 & 38.1 $\pm$ 1.7 & 45.2 $\pm$ 1.8 \\
all & 66.6 $\pm$ 0.2 & 65.9 $\pm$ 0.2 & 63.7 $\pm$ 0.2 & 34.7 $\pm$ 0.2 & 28.8 $\pm$ 0.2 & 38.7 $\pm$ 0.2 \\
jv-ID & 66.6 $\pm$ 1.7 & 65.4 $\pm$ 1.7 & 59.3 $\pm$ 1.8 & 19.0 $\pm$ 1.4 & 15.3 $\pm$ 1.3 & 11.7 $\pm$ 1.2 \\
sq-AL & 66.5 $\pm$ 1.7 & 65.1 $\pm$ 1.7 & 63.6 $\pm$ 1.7 & 35.5 $\pm$ 1.7 & 28.9 $\pm$ 1.6 & 35.1 $\pm$ 1.7 \\
he-IL & 66.2 $\pm$ 1.7 & 65.9 $\pm$ 1.7 & 64.5 $\pm$ 1.7 & 28.1 $\pm$ 1.6 & 26.6 $\pm$ 1.6 & 37.8 $\pm$ 1.7 \\
es-ES & 66.2 $\pm$ 1.7 & 64.3 $\pm$ 1.7 & 62.8 $\pm$ 1.7 & 50.4 $\pm$ 1.8 & 39.7 $\pm$ 1.8 & 47.6 $\pm$ 1.8 \\
fr-FR & 66.2 $\pm$ 1.7 & 65.1 $\pm$ 1.7 & 62.2 $\pm$ 1.7 & 47.2 $\pm$ 1.8 & 39.5 $\pm$ 1.8 & 48.6 $\pm$ 1.8 \\
bn-BD & 66.2 $\pm$ 1.7 & 66.0 $\pm$ 1.7 & 63.4 $\pm$ 1.7 & 27.3 $\pm$ 1.6 & 21.6 $\pm$ 1.5 & 36.3 $\pm$ 1.7 \\
hy-AM & 66.1 $\pm$ 1.7 & 65.8 $\pm$ 1.7 & 63.1 $\pm$ 1.7 & 34.8 $\pm$ 1.7 & 26.3 $\pm$ 1.6 & 39.0 $\pm$ 1.8 \\
mn-MN & 66.0 $\pm$ 1.7 & 65.4 $\pm$ 1.7 & 63.4 $\pm$ 1.7 & 24.3 $\pm$ 1.5 & 16.4 $\pm$ 1.3 & 33.3 $\pm$ 1.7 \\
fa-IR & 65.9 $\pm$ 1.7 & 67.3 $\pm$ 1.7 & 67.0 $\pm$ 1.7 & 38.7 $\pm$ 1.8 & 31.5 $\pm$ 1.7 & 49.6 $\pm$ 1.8 \\
sl-SL & 65.9 $\pm$ 1.7 & 65.6 $\pm$ 1.7 & 64.3 $\pm$ 1.7 & 36.3 $\pm$ 1.7 & 29.9 $\pm$ 1.6 & 38.4 $\pm$ 1.7 \\
tl-PH & 65.6 $\pm$ 1.7 & 65.6 $\pm$ 1.7 & 61.1 $\pm$ 1.8 & 34.3 $\pm$ 1.7 & 26.9 $\pm$ 1.6 & 26.9 $\pm$ 1.6 \\
hi-IN & 65.4 $\pm$ 1.7 & 64.7 $\pm$ 1.7 & 63.4 $\pm$ 1.7 & 35.1 $\pm$ 1.7 & 29.4 $\pm$ 1.6 & 42.6 $\pm$ 1.8 \\
km-KH & 65.1 $\pm$ 1.7 & 65.0 $\pm$ 1.7 & 60.5 $\pm$ 1.8 & 24.9 $\pm$ 1.6 & 34.7 $\pm$ 1.7 & 35.3 $\pm$ 1.7 \\
vi-VN & 65.0 $\pm$ 1.7 & 64.5 $\pm$ 1.7 & 64.5 $\pm$ 1.7 & 26.8 $\pm$ 1.6 & 23.9 $\pm$ 1.5 & 44.1 $\pm$ 1.8 \\
cy-GB & 64.9 $\pm$ 1.7 & 63.3 $\pm$ 1.7 & 60.1 $\pm$ 1.8 & 10.0 $\pm$ 1.1 & 8.3 $\pm$ 1.0 & 17.1 $\pm$ 1.4 \\
zh-CN & 64.8 $\pm$ 1.7 & 62.9 $\pm$ 1.7 & 60.4 $\pm$ 1.8 & 25.0 $\pm$ 1.6 & 14.1 $\pm$ 1.3 & 17.7 $\pm$ 1.4 \\
pl-PL & 64.4 $\pm$ 1.7 & 64.0 $\pm$ 1.7 & 60.9 $\pm$ 1.8 & 45.9 $\pm$ 1.8 & 39.9 $\pm$ 1.8 & 49.1 $\pm$ 1.8 \\
ar-SA & 64.1 $\pm$ 1.7 & 63.4 $\pm$ 1.7 & 61.2 $\pm$ 1.8 & 29.6 $\pm$ 1.6 & 28.7 $\pm$ 1.6 & 31.2 $\pm$ 1.7 \\
ur-PK & 64.0 $\pm$ 1.7 & 62.4 $\pm$ 1.7 & 59.0 $\pm$ 1.8 & 24.0 $\pm$ 1.5 & 19.3 $\pm$ 1.4 & 30.5 $\pm$ 1.7 \\
ta-IN & 63.8 $\pm$ 1.7 & 63.5 $\pm$ 1.7 & 61.4 $\pm$ 1.7 & 34.3 $\pm$ 1.7 & 27.9 $\pm$ 1.6 & 37.0 $\pm$ 1.7 \\
te-IN & 63.8 $\pm$ 1.7 & 65.3 $\pm$ 1.7 & 62.2 $\pm$ 1.7 & 28.3 $\pm$ 1.6 & 22.5 $\pm$ 1.5 & 36.6 $\pm$ 1.7 \\
ka-GE & 63.6 $\pm$ 1.7 & 63.5 $\pm$ 1.7 & 62.8 $\pm$ 1.7 & 32.5 $\pm$ 1.7 & 30.5 $\pm$ 1.7 & 36.8 $\pm$ 1.7 \\
am-ET & 63.4 $\pm$ 1.7 & 63.0 $\pm$ 1.7 & 59.3 $\pm$ 1.8 & 16.1 $\pm$ 1.3 & 12.0 $\pm$ 1.2 & 23.8 $\pm$ 1.5 \\
sw-KE & 63.3 $\pm$ 1.7 & 63.3 $\pm$ 1.7 & 58.5 $\pm$ 1.8 & 17.1 $\pm$ 1.4 & 15.2 $\pm$ 1.3 & 13.9 $\pm$ 1.2 \\
kn-IN & 62.8 $\pm$ 1.7 & 62.3 $\pm$ 1.7 & 59.4 $\pm$ 1.8 & 30.3 $\pm$ 1.7 & 21.7 $\pm$ 1.5 & 33.4 $\pm$ 1.7 \\
zh-TW & 61.0 $\pm$ 1.8 & 59.2 $\pm$ 1.8 & 58.0 $\pm$ 1.8 & 27.4 $\pm$ 1.6 & 15.3 $\pm$ 1.3 & 18.1 $\pm$ 1.4 \\
ja-JP & 58.3 $\pm$ 1.8 & 57.8 $\pm$ 1.8 & 55.8 $\pm$ 1.8 & 9.4 $\pm$ 1.0 & 4.2 $\pm$ 0.7 & 9.8 $\pm$ 1.1 \\
\bottomrule
\end{tabular}
}
\caption{Exact match accuracy by language for our three models using the full dataset and the zero-shot setup.}
\label{tab:all_ex_match}
\end{table*}

\begin{table*}[]
\centering
\resizebox{0.85\linewidth}{!}{%
\begin{tabular}{lcccccc}
\multicolumn{7}{c}{Intent Accuracy (\%)} \\
\toprule
 & mT5 T2T Full & mT5 Enc Full & XLM-R Full & mT5 T2T Zero & mT5 Enc Zero & XLM-R Zero \\
\midrule
en-US & 87.9 $\pm$ 1.2 & 89.0 $\pm$ 1.1 & 88.3 $\pm$ 1.2 &  & & \\
sv-SE & 87.8 $\pm$ 1.2 & 88.5 $\pm$ 1.1 & 87.9 $\pm$ 1.2 & 77.1 $\pm$ 1.5 & 76.0 $\pm$ 1.5 & 85.2 $\pm$ 1.3 \\
nb-NO & 87.6 $\pm$ 1.2 & 87.7 $\pm$ 1.2 & 87.3 $\pm$ 1.2 & 76.3 $\pm$ 1.5 & 72.8 $\pm$ 1.6 & 83.6 $\pm$ 1.3 \\
da-DK & 87.5 $\pm$ 1.2 & 88.0 $\pm$ 1.2 & 86.9 $\pm$ 1.2 & 76.8 $\pm$ 1.5 & 73.4 $\pm$ 1.6 & 83.1 $\pm$ 1.3 \\
ro-RO & 87.2 $\pm$ 1.2 & 87.0 $\pm$ 1.2 & 86.9 $\pm$ 1.2 & 73.0 $\pm$ 1.6 & 70.1 $\pm$ 1.6 & 80.8 $\pm$ 1.4 \\
nl-NL & 87.2 $\pm$ 1.2 & 87.6 $\pm$ 1.2 & 86.8 $\pm$ 1.2 & 79.9 $\pm$ 1.4 & 76.4 $\pm$ 1.5 & 82.1 $\pm$ 1.4 \\
ru-RU & 87.0 $\pm$ 1.2 & 86.8 $\pm$ 1.2 & 87.2 $\pm$ 1.2 & 76.2 $\pm$ 1.5 & 73.8 $\pm$ 1.6 & 81.3 $\pm$ 1.4 \\
id-ID & 87.0 $\pm$ 1.2 & 86.8 $\pm$ 1.2 & 87.1 $\pm$ 1.2 & 77.0 $\pm$ 1.5 & 74.1 $\pm$ 1.6 & 83.1 $\pm$ 1.3 \\
fr-FR & 86.9 $\pm$ 1.2 & 87.2 $\pm$ 1.2 & 86.3 $\pm$ 1.2 & 76.9 $\pm$ 1.5 & 74.1 $\pm$ 1.6 & 80.8 $\pm$ 1.4 \\
it-IT & 86.8 $\pm$ 1.2 & 87.6 $\pm$ 1.2 & 86.6 $\pm$ 1.2 & 72.3 $\pm$ 1.6 & 71.5 $\pm$ 1.6 & 76.4 $\pm$ 1.5 \\
ms-MY & 86.8 $\pm$ 1.2 & 86.9 $\pm$ 1.2 & 86.1 $\pm$ 1.2 & 69.9 $\pm$ 1.6 & 66.0 $\pm$ 1.7 & 76.7 $\pm$ 1.5 \\
es-ES & 86.7 $\pm$ 1.2 & 86.8 $\pm$ 1.2 & 86.9 $\pm$ 1.2 & 76.6 $\pm$ 1.5 & 75.9 $\pm$ 1.5 & 78.8 $\pm$ 1.5 \\
pt-PT & 86.7 $\pm$ 1.2 & 86.9 $\pm$ 1.2 & 86.7 $\pm$ 1.2 & 74.0 $\pm$ 1.6 & 74.5 $\pm$ 1.6 & 79.5 $\pm$ 1.5 \\
fa-IR & 86.3 $\pm$ 1.2 & 87.2 $\pm$ 1.2 & 87.0 $\pm$ 1.2 & 69.0 $\pm$ 1.7 & 66.3 $\pm$ 1.7 & 81.1 $\pm$ 1.4 \\
pl-PL & 86.3 $\pm$ 1.2 & 87.1 $\pm$ 1.2 & 85.8 $\pm$ 1.3 & 76.4 $\pm$ 1.5 & 74.1 $\pm$ 1.6 & 80.7 $\pm$ 1.4 \\
de-DE & 86.2 $\pm$ 1.2 & 86.8 $\pm$ 1.2 & 85.7 $\pm$ 1.3 & 77.3 $\pm$ 1.5 & 73.9 $\pm$ 1.6 & 77.6 $\pm$ 1.5 \\
az-AZ & 86.2 $\pm$ 1.2 & 86.4 $\pm$ 1.2 & 86.2 $\pm$ 1.2 & 57.0 $\pm$ 1.8 & 55.5 $\pm$ 1.8 & 70.9 $\pm$ 1.6 \\
tr-TR & 86.1 $\pm$ 1.2 & 87.1 $\pm$ 1.2 & 86.3 $\pm$ 1.2 & 66.5 $\pm$ 1.7 & 63.7 $\pm$ 1.7 & 78.4 $\pm$ 1.5 \\
ko-KR & 86.1 $\pm$ 1.2 & 86.4 $\pm$ 1.2 & 86.5 $\pm$ 1.2 & 60.0 $\pm$ 1.8 & 61.9 $\pm$ 1.7 & 77.0 $\pm$ 1.5 \\
af-ZA & 86.0 $\pm$ 1.2 & 86.9 $\pm$ 1.2 & 85.6 $\pm$ 1.3 & 68.5 $\pm$ 1.7 & 66.5 $\pm$ 1.7 & 71.7 $\pm$ 1.6 \\
ml-IN & 86.0 $\pm$ 1.2 & 86.5 $\pm$ 1.2 & 85.1 $\pm$ 1.3 & 60.6 $\pm$ 1.8 & 57.8 $\pm$ 1.8 & 70.1 $\pm$ 1.6 \\
sq-AL & 85.9 $\pm$ 1.3 & 86.4 $\pm$ 1.2 & 86.4 $\pm$ 1.2 & 62.9 $\pm$ 1.7 & 62.0 $\pm$ 1.7 & 67.6 $\pm$ 1.7 \\
sl-SL & 85.9 $\pm$ 1.3 & 86.8 $\pm$ 1.2 & 86.3 $\pm$ 1.2 & 61.5 $\pm$ 1.7 & 59.8 $\pm$ 1.8 & 69.5 $\pm$ 1.7 \\
el-GR & 85.8 $\pm$ 1.3 & 86.6 $\pm$ 1.2 & 86.2 $\pm$ 1.2 & 71.9 $\pm$ 1.6 & 69.8 $\pm$ 1.6 & 74.0 $\pm$ 1.6 \\
vi-VN & 85.8 $\pm$ 1.3 & 87.2 $\pm$ 1.2 & 86.3 $\pm$ 1.2 & 64.2 $\pm$ 1.7 & 62.7 $\pm$ 1.7 & 79.2 $\pm$ 1.5 \\
hi-IN & 85.6 $\pm$ 1.3 & 86.2 $\pm$ 1.2 & 85.8 $\pm$ 1.3 & 62.4 $\pm$ 1.7 & 59.3 $\pm$ 1.8 & 74.8 $\pm$ 1.6 \\
hu-HU & 85.4 $\pm$ 1.3 & 86.9 $\pm$ 1.2 & 86.2 $\pm$ 1.2 & 68.0 $\pm$ 1.7 & 66.4 $\pm$ 1.7 & 77.1 $\pm$ 1.5 \\
all & 85.3 $\pm$ 0.2 & 86.1 $\pm$ 0.2 & 85.1 $\pm$ 0.2 & 62.9 $\pm$ 0.2 & 61.2 $\pm$ 0.2 & 70.6 $\pm$ 0.2 \\
is-IS & 85.3 $\pm$ 1.3 & 85.9 $\pm$ 1.3 & 85.3 $\pm$ 1.3 & 59.0 $\pm$ 1.8 & 55.9 $\pm$ 1.8 & 66.7 $\pm$ 1.7 \\
fi-FI & 85.3 $\pm$ 1.3 & 86.7 $\pm$ 1.2 & 85.5 $\pm$ 1.3 & 69.7 $\pm$ 1.7 & 68.5 $\pm$ 1.7 & 80.2 $\pm$ 1.4 \\
zh-CN & 85.2 $\pm$ 1.3 & 85.8 $\pm$ 1.3 & 84.9 $\pm$ 1.3 & 55.7 $\pm$ 1.8 & 51.6 $\pm$ 1.8 & 61.9 $\pm$ 1.7 \\
lv-LV & 85.2 $\pm$ 1.3 & 86.6 $\pm$ 1.2 & 86.1 $\pm$ 1.2 & 61.0 $\pm$ 1.8 & 60.0 $\pm$ 1.8 & 69.2 $\pm$ 1.7 \\
th-TH & 85.2 $\pm$ 1.3 & 85.5 $\pm$ 1.3 & 84.7 $\pm$ 1.3 & 72.8 $\pm$ 1.6 & 69.6 $\pm$ 1.7 & 77.4 $\pm$ 1.5 \\
tl-PH & 85.1 $\pm$ 1.3 & 87.0 $\pm$ 1.2 & 84.6 $\pm$ 1.3 & 64.9 $\pm$ 1.7 & 63.8 $\pm$ 1.7 & 63.7 $\pm$ 1.7 \\
mn-MN & 84.9 $\pm$ 1.3 & 86.0 $\pm$ 1.2 & 84.3 $\pm$ 1.3 & 47.8 $\pm$ 1.8 & 47.2 $\pm$ 1.8 & 64.4 $\pm$ 1.7 \\
kn-IN & 84.9 $\pm$ 1.3 & 84.9 $\pm$ 1.3 & 84.0 $\pm$ 1.3 & 56.7 $\pm$ 1.8 & 51.8 $\pm$ 1.8 & 63.5 $\pm$ 1.7 \\
te-IN & 84.9 $\pm$ 1.3 & 85.5 $\pm$ 1.3 & 84.5 $\pm$ 1.3 & 55.0 $\pm$ 1.8 & 52.2 $\pm$ 1.8 & 68.2 $\pm$ 1.7 \\
bn-BD & 84.8 $\pm$ 1.3 & 85.8 $\pm$ 1.3 & 84.1 $\pm$ 1.3 & 56.5 $\pm$ 1.8 & 52.1 $\pm$ 1.8 & 66.0 $\pm$ 1.7 \\
he-IL & 84.6 $\pm$ 1.3 & 86.2 $\pm$ 1.2 & 85.9 $\pm$ 1.3 & 64.7 $\pm$ 1.7 & 64.0 $\pm$ 1.7 & 73.2 $\pm$ 1.6 \\
my-MM & 84.6 $\pm$ 1.3 & 85.2 $\pm$ 1.3 & 83.6 $\pm$ 1.3 & 58.3 $\pm$ 1.8 & 58.4 $\pm$ 1.8 & 67.6 $\pm$ 1.7 \\
jv-ID & 84.5 $\pm$ 1.3 & 85.3 $\pm$ 1.3 & 82.9 $\pm$ 1.4 & 47.6 $\pm$ 1.8 & 49.3 $\pm$ 1.8 & 46.5 $\pm$ 1.8 \\
hy-AM & 84.5 $\pm$ 1.3 & 85.6 $\pm$ 1.3 & 84.4 $\pm$ 1.3 & 63.8 $\pm$ 1.7 & 62.2 $\pm$ 1.7 & 71.6 $\pm$ 1.6 \\
ta-IN & 84.4 $\pm$ 1.3 & 85.2 $\pm$ 1.3 & 83.5 $\pm$ 1.3 & 61.3 $\pm$ 1.8 & 58.0 $\pm$ 1.8 & 68.1 $\pm$ 1.7 \\
ur-PK & 84.3 $\pm$ 1.3 & 85.1 $\pm$ 1.3 & 83.2 $\pm$ 1.3 & 47.2 $\pm$ 1.8 & 49.0 $\pm$ 1.8 & 65.6 $\pm$ 1.7 \\
sw-KE & 84.0 $\pm$ 1.3 & 85.8 $\pm$ 1.3 & 83.1 $\pm$ 1.3 & 45.6 $\pm$ 1.8 & 44.7 $\pm$ 1.8 & 46.6 $\pm$ 1.8 \\
cy-GB & 83.7 $\pm$ 1.3 & 84.9 $\pm$ 1.3 & 82.6 $\pm$ 1.4 & 29.6 $\pm$ 1.6 & 33.1 $\pm$ 1.7 & 46.9 $\pm$ 1.8 \\
ja-JP & 83.5 $\pm$ 1.3 & 85.8 $\pm$ 1.3 & 83.9 $\pm$ 1.3 & 25.7 $\pm$ 1.6 & 27.1 $\pm$ 1.6 & 44.8 $\pm$ 1.8 \\
zh-TW & 82.9 $\pm$ 1.4 & 83.8 $\pm$ 1.3 & 83.0 $\pm$ 1.3 & 56.1 $\pm$ 1.8 & 52.2 $\pm$ 1.8 & 60.4 $\pm$ 1.8 \\
am-ET & 82.7 $\pm$ 1.4 & 84.2 $\pm$ 1.3 & 81.7 $\pm$ 1.4 & 36.8 $\pm$ 1.7 & 36.6 $\pm$ 1.7 & 51.9 $\pm$ 1.8 \\
ar-SA & 81.8 $\pm$ 1.4 & 82.2 $\pm$ 1.4 & 80.7 $\pm$ 1.4 & 59.0 $\pm$ 1.8 & 58.8 $\pm$ 1.8 & 62.8 $\pm$ 1.7 \\
ka-GE & 79.9 $\pm$ 1.4 & 81.3 $\pm$ 1.4 & 80.3 $\pm$ 1.4 & 59.3 $\pm$ 1.8 & 58.4 $\pm$ 1.8 & 61.2 $\pm$ 1.8 \\
km-KH & 79.0 $\pm$ 1.5 & 79.1 $\pm$ 1.5 & 77.2 $\pm$ 1.5 & 60.2 $\pm$ 1.8 & 58.7 $\pm$ 1.8 & 61.3 $\pm$ 1.8 \\

\bottomrule
\end{tabular}
}
\caption{Intent accuracy by language for our three models using the full dataset and the zero-shot setup.}
\label{tab:all_intent_acc}
\end{table*}

\begin{table*}[]
\centering
\resizebox{0.85\linewidth}{!}{%
\begin{tabular}{lcccccc}
\multicolumn{7}{c}{Micro-Averaged Slot F1 (\%)} \\
\toprule
 & mT5 T2T Full & mT5 Enc Full & XLM-R Full & mT5 T2T Zero & mT5 Enc Zero & XLM-R Zero \\
\midrule
th-TH & 86.8 $\pm$ 0.7 & 85.7 $\pm$ 0.7 & 83.5 $\pm$ 0.7 & 34.5 $\pm$ 0.9 & 59.5 $\pm$ 1.0 & 57.4 $\pm$ 1.0 \\
my-MM & 82.2 $\pm$ 0.7 & 82.1 $\pm$ 0.7 & 79.0 $\pm$ 0.7 & 26.0 $\pm$ 0.8 & 38.0 $\pm$ 0.9 & 48.9 $\pm$ 0.9 \\
en-US & 81.6 $\pm$ 0.5 & 80.4 $\pm$ 0.5 & 78.7 $\pm$ 0.6 &  & & \\
km-KH & 81.0 $\pm$ 0.8 & 81.9 $\pm$ 0.8 & 77.9 $\pm$ 0.8 & 27.9 $\pm$ 0.9 & 58.2 $\pm$ 1.0 & 53.6 $\pm$ 1.0 \\
sv-SE & 80.9 $\pm$ 0.6 & 79.6 $\pm$ 0.6 & 78.5 $\pm$ 0.6 & 64.2 $\pm$ 0.7 & 56.8 $\pm$ 0.7 & 68.4 $\pm$ 0.7 \\
nb-NO & 80.0 $\pm$ 0.6 & 77.8 $\pm$ 0.6 & 76.0 $\pm$ 0.6 & 58.8 $\pm$ 0.7 & 56.0 $\pm$ 0.7 & 65.1 $\pm$ 0.7 \\
ko-KR & 79.6 $\pm$ 0.7 & 78.9 $\pm$ 0.7 & 77.8 $\pm$ 0.7 & 46.8 $\pm$ 0.8 & 36.0 $\pm$ 0.8 & 56.0 $\pm$ 0.8 \\
da-DK & 79.4 $\pm$ 0.6 & 79.1 $\pm$ 0.6 & 77.7 $\pm$ 0.6 & 58.5 $\pm$ 0.7 & 54.6 $\pm$ 0.7 & 64.6 $\pm$ 0.7 \\
fi-FI & 79.4 $\pm$ 0.7 & 79.2 $\pm$ 0.7 & 77.2 $\pm$ 0.7 & 49.1 $\pm$ 0.8 & 48.9 $\pm$ 0.8 & 62.1 $\pm$ 0.8 \\
de-DE & 78.8 $\pm$ 0.6 & 78.6 $\pm$ 0.6 & 76.2 $\pm$ 0.6 & 64.3 $\pm$ 0.7 & 55.6 $\pm$ 0.7 & 60.0 $\pm$ 0.7 \\
ru-RU & 78.7 $\pm$ 0.6 & 76.3 $\pm$ 0.6 & 74.9 $\pm$ 0.6 & 61.6 $\pm$ 0.7 & 55.4 $\pm$ 0.7 & 63.3 $\pm$ 0.7 \\
ms-MY & 78.4 $\pm$ 0.6 & 77.4 $\pm$ 0.6 & 75.5 $\pm$ 0.6 & 51.5 $\pm$ 0.7 & 48.2 $\pm$ 0.7 & 55.9 $\pm$ 0.7 \\
af-ZA & 78.3 $\pm$ 0.6 & 76.5 $\pm$ 0.6 & 74.6 $\pm$ 0.6 & 51.9 $\pm$ 0.7 & 52.3 $\pm$ 0.7 & 57.3 $\pm$ 0.7 \\
is-IS & 78.2 $\pm$ 0.6 & 77.7 $\pm$ 0.6 & 75.2 $\pm$ 0.6 & 39.3 $\pm$ 0.7 & 37.9 $\pm$ 0.7 & 45.2 $\pm$ 0.7 \\
nl-NL & 78.1 $\pm$ 0.6 & 76.5 $\pm$ 0.6 & 75.5 $\pm$ 0.6 & 61.6 $\pm$ 0.7 & 54.3 $\pm$ 0.7 & 62.4 $\pm$ 0.7 \\
jv-ID & 78.1 $\pm$ 0.6 & 76.1 $\pm$ 0.6 & 70.9 $\pm$ 0.7 & 29.6 $\pm$ 0.7 & 26.7 $\pm$ 0.7 & 24.7 $\pm$ 0.6 \\
hu-HU & 78.0 $\pm$ 0.6 & 77.5 $\pm$ 0.6 & 75.3 $\pm$ 0.6 & 46.1 $\pm$ 0.7 & 45.8 $\pm$ 0.7 & 56.8 $\pm$ 0.7 \\
tr-TR & 77.9 $\pm$ 0.6 & 76.1 $\pm$ 0.7 & 74.9 $\pm$ 0.7 & 48.8 $\pm$ 0.8 & 41.9 $\pm$ 0.8 & 52.8 $\pm$ 0.8 \\
lv-LV & 77.8 $\pm$ 0.6 & 77.1 $\pm$ 0.6 & 76.3 $\pm$ 0.6 & 47.2 $\pm$ 0.8 & 41.6 $\pm$ 0.7 & 53.0 $\pm$ 0.8 \\
ka-GE & 77.6 $\pm$ 0.7 & 77.1 $\pm$ 0.7 & 76.8 $\pm$ 0.7 & 43.5 $\pm$ 0.9 & 48.6 $\pm$ 0.9 & 55.9 $\pm$ 0.9 \\
ro-RO & 77.6 $\pm$ 0.6 & 74.1 $\pm$ 0.6 & 72.4 $\pm$ 0.6 & 56.3 $\pm$ 0.7 & 48.6 $\pm$ 0.7 & 60.8 $\pm$ 0.7 \\
el-GR & 77.0 $\pm$ 0.6 & 75.5 $\pm$ 0.6 & 73.4 $\pm$ 0.6 & 54.8 $\pm$ 0.7 & 51.7 $\pm$ 0.7 & 54.4 $\pm$ 0.7 \\
id-ID & 76.9 $\pm$ 0.6 & 75.6 $\pm$ 0.6 & 73.6 $\pm$ 0.6 & 55.6 $\pm$ 0.7 & 51.0 $\pm$ 0.7 & 59.7 $\pm$ 0.7 \\
all & 76.8 $\pm$ 0.1 & 75.4 $\pm$ 0.1 & 73.6 $\pm$ 0.1 & 44.8 $\pm$ 0.1 & 41.6 $\pm$ 0.1 & 50.3 $\pm$ 0.1 \\
az-AZ & 76.8 $\pm$ 0.6 & 75.6 $\pm$ 0.7 & 74.1 $\pm$ 0.7 & 40.4 $\pm$ 0.7 & 33.8 $\pm$ 0.7 & 46.6 $\pm$ 0.8 \\
he-IL & 76.7 $\pm$ 0.6 & 75.1 $\pm$ 0.7 & 74.0 $\pm$ 0.7 & 30.6 $\pm$ 0.7 & 35.5 $\pm$ 0.7 & 49.3 $\pm$ 0.8 \\
pt-PT & 76.6 $\pm$ 0.6 & 74.9 $\pm$ 0.6 & 73.3 $\pm$ 0.6 & 56.3 $\pm$ 0.7 & 46.6 $\pm$ 0.7 & 58.2 $\pm$ 0.7 \\
ml-IN & 76.6 $\pm$ 0.7 & 76.1 $\pm$ 0.7 & 74.8 $\pm$ 0.7 & 42.1 $\pm$ 0.8 & 45.5 $\pm$ 0.8 & 52.5 $\pm$ 0.8 \\
it-IT & 76.4 $\pm$ 0.6 & 73.7 $\pm$ 0.6 & 72.3 $\pm$ 0.6 & 58.7 $\pm$ 0.7 & 50.0 $\pm$ 0.7 & 57.3 $\pm$ 0.7 \\
bn-BD & 76.4 $\pm$ 0.6 & 75.1 $\pm$ 0.6 & 73.4 $\pm$ 0.6 & 39.6 $\pm$ 0.7 & 37.2 $\pm$ 0.7 & 52.3 $\pm$ 0.7 \\
cy-GB & 76.3 $\pm$ 0.6 & 73.5 $\pm$ 0.6 & 71.2 $\pm$ 0.6 & 21.8 $\pm$ 0.6 & 21.5 $\pm$ 0.5 & 30.1 $\pm$ 0.6 \\
sq-AL & 75.9 $\pm$ 0.6 & 73.7 $\pm$ 0.6 & 72.0 $\pm$ 0.6 & 48.3 $\pm$ 0.7 & 41.9 $\pm$ 0.7 & 50.0 $\pm$ 0.7 \\
tl-PH & 75.8 $\pm$ 0.6 & 74.6 $\pm$ 0.6 & 71.6 $\pm$ 0.6 & 44.7 $\pm$ 0.6 & 37.1 $\pm$ 0.6 & 36.1 $\pm$ 0.6 \\
mn-MN & 75.8 $\pm$ 0.6 & 74.1 $\pm$ 0.6 & 73.7 $\pm$ 0.7 & 36.6 $\pm$ 0.7 & 26.9 $\pm$ 0.7 & 45.0 $\pm$ 0.7 \\
ar-SA & 75.7 $\pm$ 0.7 & 75.4 $\pm$ 0.7 & 73.8 $\pm$ 0.7 & 39.7 $\pm$ 0.8 & 44.6 $\pm$ 0.8 & 48.4 $\pm$ 0.8 \\
fr-FR & 75.6 $\pm$ 0.6 & 73.5 $\pm$ 0.6 & 70.9 $\pm$ 0.6 & 54.2 $\pm$ 0.7 & 51.2 $\pm$ 0.7 & 59.1 $\pm$ 0.7 \\
es-ES & 75.5 $\pm$ 0.6 & 72.8 $\pm$ 0.6 & 71.0 $\pm$ 0.6 & 61.1 $\pm$ 0.7 & 50.4 $\pm$ 0.7 & 57.1 $\pm$ 0.7 \\
fa-IR & 75.4 $\pm$ 0.6 & 76.6 $\pm$ 0.6 & 76.6 $\pm$ 0.6 & 49.4 $\pm$ 0.7 & 46.9 $\pm$ 0.7 & 60.2 $\pm$ 0.6 \\
sl-SL & 75.4 $\pm$ 0.6 & 74.3 $\pm$ 0.6 & 72.2 $\pm$ 0.7 & 49.0 $\pm$ 0.7 & 45.6 $\pm$ 0.7 & 53.1 $\pm$ 0.7 \\
hy-AM & 75.3 $\pm$ 0.7 & 74.1 $\pm$ 0.7 & 72.4 $\pm$ 0.7 & 41.7 $\pm$ 0.7 & 39.1 $\pm$ 0.7 & 50.0 $\pm$ 0.8 \\
hi-IN & 75.0 $\pm$ 0.6 & 73.5 $\pm$ 0.6 & 72.3 $\pm$ 0.6 & 49.6 $\pm$ 0.7 & 45.1 $\pm$ 0.7 & 54.6 $\pm$ 0.7 \\
zh-CN & 74.5 $\pm$ 0.5 & 71.2 $\pm$ 0.5 & 70.0 $\pm$ 0.5 & 33.4 $\pm$ 0.5 & 20.9 $\pm$ 0.5 & 24.8 $\pm$ 0.5 \\
ta-IN & 74.3 $\pm$ 0.7 & 72.6 $\pm$ 0.7 & 71.8 $\pm$ 0.7 & 45.8 $\pm$ 0.8 & 45.9 $\pm$ 0.8 & 50.3 $\pm$ 0.8 \\
vi-VN & 74.2 $\pm$ 0.5 & 72.3 $\pm$ 0.5 & 73.3 $\pm$ 0.5 & 28.8 $\pm$ 0.5 & 36.0 $\pm$ 0.6 & 53.9 $\pm$ 0.6 \\
am-ET & 73.8 $\pm$ 0.7 & 73.7 $\pm$ 0.7 & 70.0 $\pm$ 0.7 & 25.9 $\pm$ 0.7 & 21.3 $\pm$ 0.6 & 39.0 $\pm$ 0.8 \\
sw-KE & 73.8 $\pm$ 0.6 & 72.9 $\pm$ 0.6 & 68.7 $\pm$ 0.7 & 25.9 $\pm$ 0.6 & 28.2 $\pm$ 0.6 & 27.7 $\pm$ 0.6 \\
te-IN & 73.0 $\pm$ 0.7 & 74.7 $\pm$ 0.7 & 71.4 $\pm$ 0.7 & 41.1 $\pm$ 0.7 & 39.4 $\pm$ 0.7 & 51.6 $\pm$ 0.7 \\
ur-PK & 73.0 $\pm$ 0.6 & 71.2 $\pm$ 0.6 & 68.0 $\pm$ 0.6 & 40.1 $\pm$ 0.6 & 32.6 $\pm$ 0.6 & 41.4 $\pm$ 0.6 \\
zh-TW & 72.9 $\pm$ 0.5 & 68.8 $\pm$ 0.6 & 68.7 $\pm$ 0.6 & 34.4 $\pm$ 0.6 & 22.6 $\pm$ 0.5 & 25.2 $\pm$ 0.5 \\
pl-PL & 72.9 $\pm$ 0.7 & 71.7 $\pm$ 0.7 & 69.0 $\pm$ 0.7 & 53.4 $\pm$ 0.7 & 49.3 $\pm$ 0.7 & 58.0 $\pm$ 0.7 \\
kn-IN & 72.2 $\pm$ 0.7 & 71.3 $\pm$ 0.7 & 69.2 $\pm$ 0.7 & 40.4 $\pm$ 0.8 & 38.3 $\pm$ 0.8 & 47.8 $\pm$ 0.8 \\
ja-JP & 67.6 $\pm$ 0.4 & 64.5 $\pm$ 0.4 & 63.3 $\pm$ 0.4 & 13.9 $\pm$ 0.3 & 6.3 $\pm$ 0.2 & 15.4 $\pm$ 0.3 \\
\bottomrule
\end{tabular}
}
\caption{Micro-averaged slot-filling F1 by language for our three models using the full dataset and the zero-shot setup.}
\label{tab:all_slot_f1}
\end{table*}

\section{A summary of model performance on language characteristics}
\label{sect:perf_lang_char}
\begin{figure*}
\def\tabularxcolumn#1{m{#1}}
\begin{tabular}{c}
\captionsetup[subfloat]{labelformat=empty}
\subfloat[]{\includegraphics[width=1.0\textwidth]{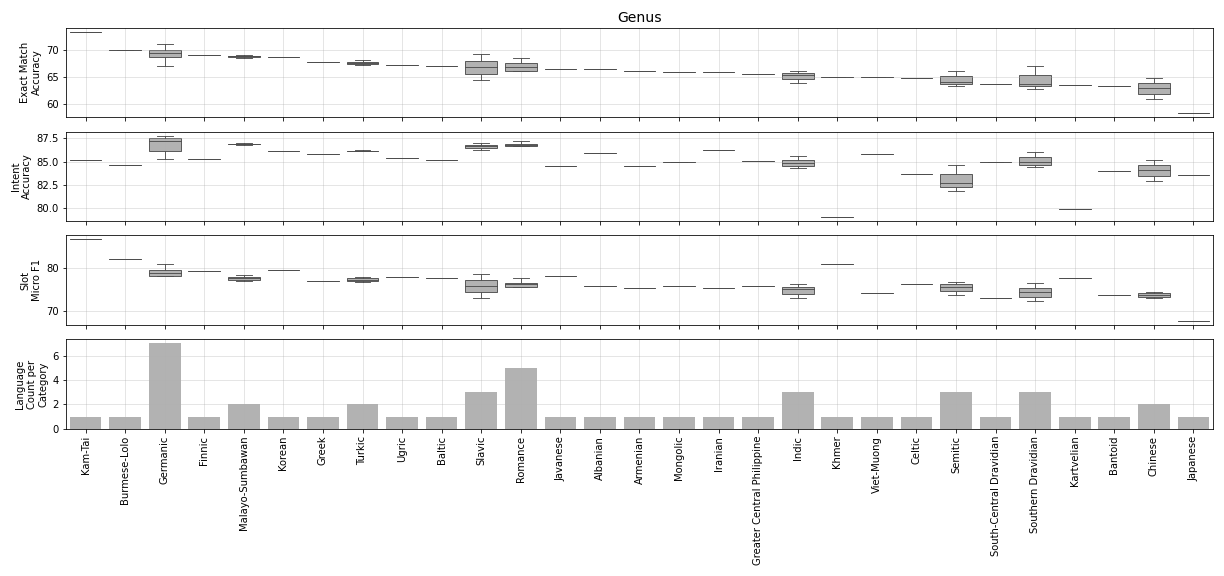}} \\
\captionsetup[subfloat]{labelformat=empty}
\subfloat[]{\includegraphics[width=1.0\textwidth]{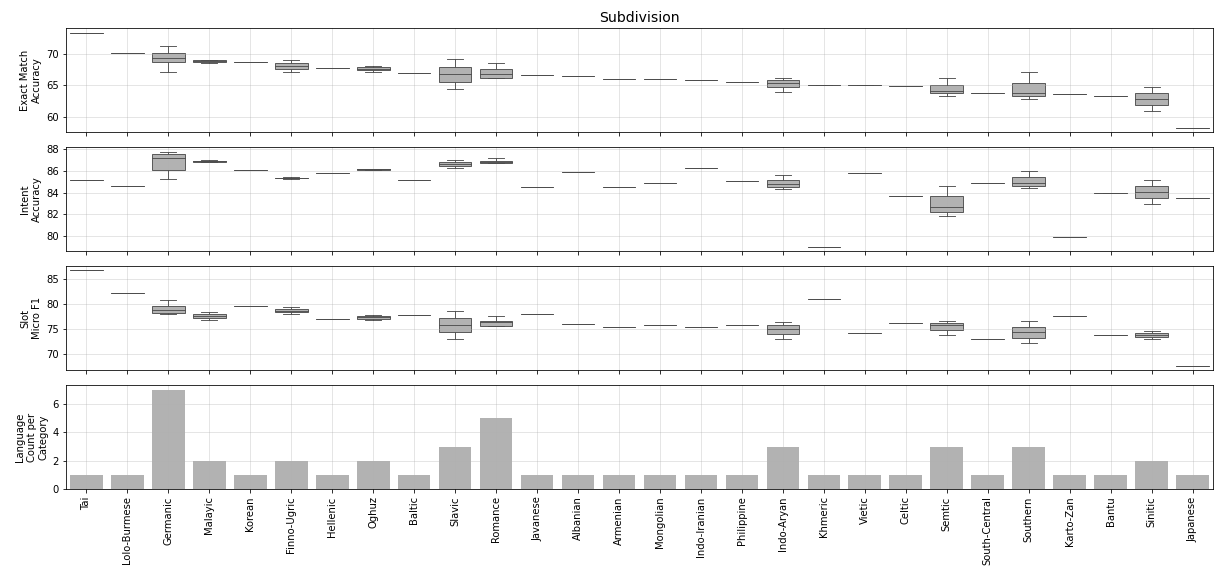}} \\
\end{tabular}
\caption{mT5 Text-to-Text performance grouped by Genus and Subdivision. The categories of each language characteristic are sorted by exact match accuracy for readability. The number of languages falling into each category is provided in the bar chart in the lowest panel for each characteristic.}
\label{genus_subdivision}
\end{figure*}

\begin{figure*}
\def\tabularxcolumn#1{m{#1}}
\begin{tabular}{c}
\captionsetup[subfloat]{labelformat=empty}
\subfloat[]{\includegraphics[width=1.0\textwidth]{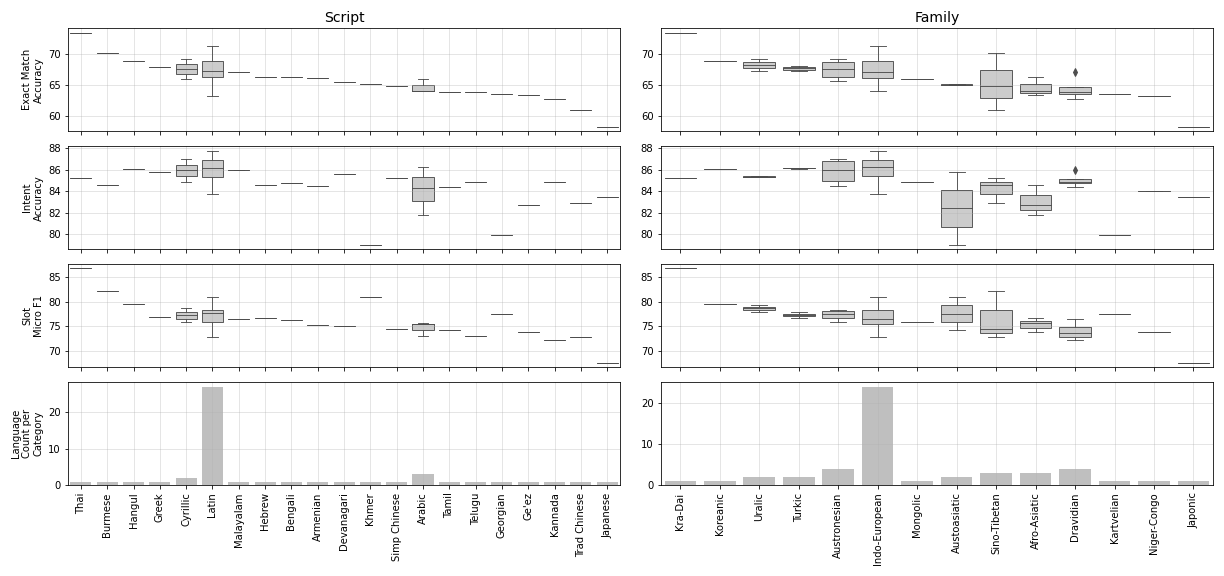}} \\
\captionsetup[subfloat]{labelformat=empty}
\subfloat[]{\includegraphics[width=1.0\textwidth]{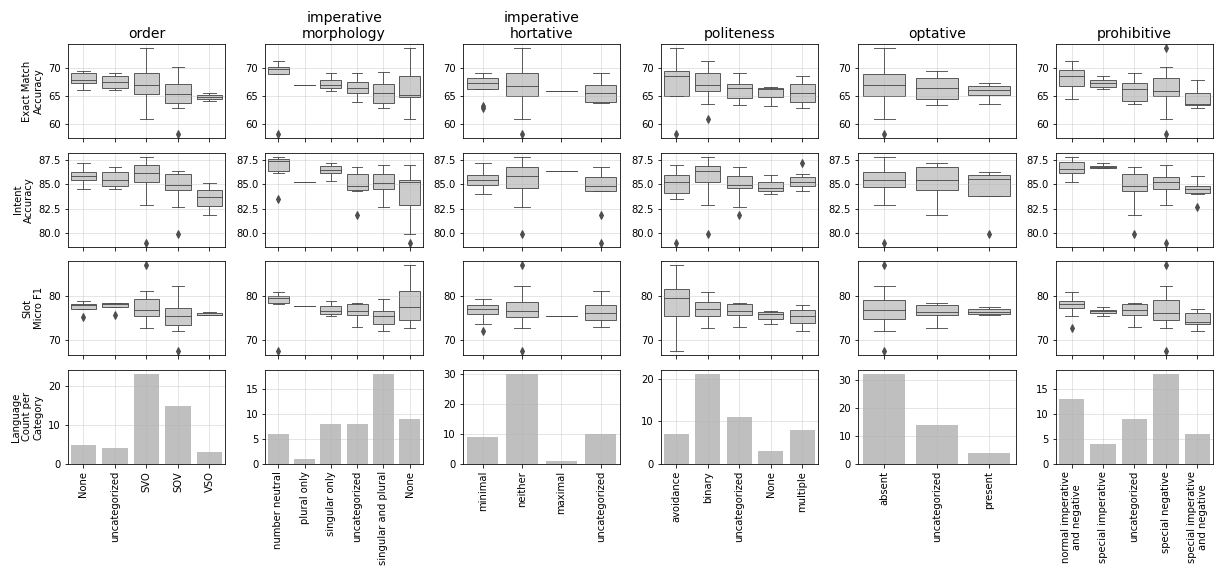}} \\
\end{tabular}
\caption{mT5 Text-to-Text performance grouped by Script, Family, Order, Politeness, Imperative Morphology, Imperative Hortative, Optative and Prohibitive. As with Figure~\ref{genus_subdivision}, the categories of each language characteristic are sorted by exact match accuracy for readability. The number of languages falling into each category is provided in the bar chart in the lowest panel for each characteristic.}
\label{other_perf}
\end{figure*}

We pick our best performing model, mT5 Text-to-Text, and provide a summary of its performance on different language characteristics in Figures~\ref{genus_subdivision} and~\ref{other_perf}.

\end{document}